\documentclass[journal]{IEEEtai}
\usepackage{url}

\usepackage[hidelinks]{hyperref}
\usepackage{latexsym}
\usepackage{amssymb}
\usepackage{amsmath}
\usepackage{amsthm}
\usepackage{booktabs}
\let\labelindent\relax
\usepackage{enumitem}
\usepackage{graphicx}
\usepackage[square,numbers]{natbib}
\usepackage{microtype}      
\usepackage{xcolor}         
\usepackage{url} 
\usepackage{amsmath}
\usepackage{amssymb}
\usepackage{xcolor}

\usepackage{pgfplots}
\usepackage{multirow}
\usepackage{subcaption}
\pgfplotsset{compat=newest}

\usepackage{color,array}
\usepackage{graphicx}

\newtheorem{theorem}{Theorem}
\newtheorem{lemma}{Lemma}
\newtheorem{corollary}[theorem]{Corollary}
\newtheorem{proposition}[theorem]{Proposition}

\begin{document}

\title{Selective Prior Synchronization via SYNC Loss}

\author{
Ishan Mishra\IEEEauthorrefmark{1}\IEEEauthorrefmark{2},
Jiajie Li\IEEEauthorrefmark{1}\IEEEauthorrefmark{3},
Deepak Mishra\IEEEauthorrefmark{2},
and Jinjun Xiong\IEEEauthorrefmark{3}%
\thanks{\IEEEauthorrefmark{1}Equal Contribution.}
\thanks{\IEEEauthorrefmark{2}Indian Institute of Technology Jodhpur, India.}%
\thanks{\IEEEauthorrefmark{3}State University of New York at Buffalo, USA.}%
}

\markboth{}
{Mishra \MakeLowercase{\textit{et al.}}: Selective Prior Synchronization via SYNC Loss}

\maketitle

\begin{abstract}
Prediction under uncertainty is a critical requirement for the deep neural network to succeed responsibly.
This paper focuses on selective prediction, which allows DNNs to make informed decisions about when to predict or abstain based on the uncertainty level of their predictions.
Current methods are either ad-hoc such as SelectiveNet, focusing on how to modify the network architecture or objective function, or post-hoc such as softmax response, achieving selective prediction through analyzing the model's probabilistic outputs. 
We observe that post-hoc methods implicitly generate uncertainty information, termed the \textit{selective prior}, which has traditionally been used only during inference. We argue that the \textit{selective prior} provided by the selection mechanism is equally vital during the training stage.
Therefore, we propose the SYNC loss which introduces a novel integration of ad-hoc and post-hoc method. Specifically, our approach incorporates the softmax response into the training process of SelectiveNet, enhancing its selective prediction capabilities by examining the selective prior.
Evaluated across various datasets, including CIFAR-100, ImageNet-100, and Stanford Cars, our method not only enhances the model's generalization capabilities but also surpasses previous works in selective prediction performance, and sets new benchmarks for state-of-the-art performance.
\end{abstract}



\section{Introduction}
Despite the recent significant development of deep neural networks (DNNs) ~\citep{he2016deep} ~\citep{zhang2018shufflenet} ~\citep{tan2019efficientnet} with amazing results on various published datasets, their adoption in real-world mission-critical applications still faces some
major challenges. One of the most critical challenges
is  DNN models' lack of self-awareness and the tendency to fail silently~\citep{holzinger2017we}. 
This is in direct contrast to human intelligence. Humans are more aware of the uncertainty in their predictions, which makes them more cautious 
in making hasty decisions.
For example, in a medical setting, whenever doctors are in doubt about a diagnosis, 
the right course of action is not to rush to a conclusion but to conduct further investigations.
Similarly, in a driving setting, human drivers would slow down whenever they cannot  
recognize an object in front of their vehicles. Therefore,
to entrust DNNs with more and more mission-critical
decisions, we need to equip DNNs 
with an awareness of their task competency, or equivalently,
a reject option for their decision-making, which is known as the Selective Prediction problem.

Recent studies have explored equipping neural networks with the ability to abstain from making uncertain predictions~\citep{bartlett2008classification,cortes2016boosting,charoenphakdee2021classification}. These methods rely on various metrics to measure model confidence and determine whether to accept or reject a prediction.

We broadly categorize these approaches based on whether they modify the model's training procedure or not:

Post-hoc Methods analyze confidence using outputs from pre-trained models without modifying the training phase. For example, \cite{geifman2017selective} proposed using the maximal activation of a classification network’s softmax layer as a confidence measure. Predictions below a manually chosen threshold are rejected. Similarly, MC-Dropout~\citep{gal2016dropout} applies dropout at inference time, performing multiple forward passes to generate a predictive distribution, thus quantifying uncertainty. The advantage of post-hoc methods is their ease of use, as they can be applied directly to any pre-trained network. However, they do not explicitly enhance the network's intrinsic selective prediction capabilities; instead, they only assess existing model uncertainty.

Ad-hoc Methods modify either the network architecture or the training procedure itself. For instance, Deep Gamblers (DG)\citep{liu2019deep} introduces an additional ``gambling" class to handle uncertain predictions explicitly. SelectiveNet (SN)\cite{geifman2019selectivenet}, on the other hand, integrates a dedicated selection head directly predicting confidence to determine abstention decisions. This selection head and the primary model are trained jointly, optimizing both accuracy and coverage under predefined confidence levels. While ad-hoc methods can improve generalization, recent findings by \cite{feng2022towards} indicate that such approaches primarily provide the network with mechanisms to express uncertainty explicitly. They do not inherently teach the network to consciously recognize whether a given sample is intrinsically challenging or easy to classify.

Therefore, in this paper, we propose combining the complementary strengths of post-hoc and ad-hoc selective prediction methods. Specifically, we leverage the implicit uncertainty estimates—referred to as the selective prior—from post-hoc methods (e.g., Softmax Response) and explicitly incorporate this knowledge into the training stage of ad-hoc methods (e.g., SelectiveNet). By doing so, we train the model to align its selective prediction uncertainty (learned explicitly through ad-hoc training) with the model's inherent uncertainty (implicitly present in post-hoc outputs), ensuring these two estimates remain synchronized. Our approach, which we term SYNC loss, is simple, effective, and scalable, equipping any deep neural network with robust selective prediction capabilities.
The contributions of this paper are:
\begin{itemize}
    \item We propose a novel loss function designed to integrate prior knowledge effectively into the training process of neural networks. By incorporating prior knowledge, the network gains awareness of uncertainty during training, thereby improving its selective prediction capabilities.
    \item We introduce a new score function specifically tailored to calibrate the selective score of the model. This novel score function further enhances the selective prediction accuracy and overall generalizability. 
    \item We provide a formal theoretical analysis of our proposed loss and scoring functions, establishing key properties such as Lipschitz continuity of our Softmax-Power (SMP) score and global smoothness of the SYNC loss objective. This analysis ensures stable optimization, convergence, and robustness of the proposed approach.
    \item We validate our approach on CIFAR-100, Stanford Cars and ImageNet-100 dataset. This demonstrates the scalability and generalizability of our methods across diverse domains. We observe that our approach outperforms the existing state-of-the-art baselines such as SN and DG on various datasets, demonstrating its efficacy.

\end{itemize}

\section{Background}
In this section, we define Selective Prediction and outline key methods in Selective Classification.
Selective Classification methods are categorized into post-hoc and ad-hoc, based on the need for model and training modifications.
We start with a formal definition, and we then review leading approaches, providing a concise overview of this field's essentials.

\paragraph{Selective prediction}
We formally define the problem as follows. Let $\mathcal{X}$ be the input space (such as images) and $\mathcal{Y}$ be the label space (such as the class labels for the classification problem or the numerical outputs for the regression problems), 
and $P(X , Y)$ represents the data distribution over $\mathcal{X}\times \mathcal{Y}$. 
A selective prediction model~\cite{el2010foundations} is a pair of functions of ($f,g$), 
where $f:\mathcal{X}\rightarrow\mathcal{Y}$ is a {\em prediction function} and 
$g: \mathcal{X}\rightarrow\{0,1\}$ is a binary {\em selection function}. Their relationship is given by
\begin{equation}
(f,g)(x) \triangleq\left\{
\begin{aligned}
& f(x) , & g(x)=1; \\
& \text{abstain} , &g(x)=0. 
\end{aligned}
\right.
\end{equation}
\noindent
The selective prediction model offers model prediction $f(x)$ when the selection function
$g(x)=1$. Otherwise, the model abstains from prediction. 
Therefore, the selection function $g(x)$ is used to indicate whether the model should predict or refrain from making the prediction.

In prior works~\citep{geifman2019selectivenet,feng2022towards,liu2019deep}, the selection function $g(x)$ is defined by comparing a regression function $r(x)$ against a threshold $\tau$, i.e., 
\begin{equation}
\label{gx}
g(x)=\left\{
\begin{aligned}
& 1 , & r(x) > \tau; \\
& 0 , & otherwise.
\end{aligned}
\right.
\end{equation}
The regression function, referred to as the selective score $r(x)\in [0, 1]$, serves to quantify the model's confidence level. By establishing a threshold $\tau$, predictions deemed risky can be filtered out.


The performance of a selective prediction model can be measured through {\em coverage} and {\em selective risk}. Coverage is defined as the probability mass of the subset of regions in $\mathcal{X}$ where the model offers prediction, i.e.,
$ c = \mathbb{E}_P[g(x)]$.
Given a loss function $\ell: \mathcal{Y}\times\mathcal{Y}\rightarrow \mathbb{R}^+$ that
measures the differences between what the model offers to predict and the true class label,
the selective risk of 
 ($f,g$) is defined as
\begin{equation}
R(f,g) = \frac{\mathbb{E}_P[\ell(f(x), y)g(x)]}{c}.
\end{equation}

Formally, given a target coverage rate \( \overline{c}\), an optimal selective model
\( (f_\theta, g_\theta) \) parameterized by \( \theta \) can be found by solving the following optimization problem:
\begin{equation}
\theta^* = \arg\min_{\theta} R(f_\theta (x), g_\theta (x)), \text{ s.t. } \mathbb{E}_P \left[g (x)\right] \geq \overline{c}.
\label{eqn_opt}
\end{equation}
Several studies have highlighted the application of selective prediction across various domains. Works by~\cite{NEURIPS2021_2cb6b103},~\cite{mohri2023learning},~\cite{7344808}, and~\cite{charoenphakdee2021classification} demonstrate its versatility, while studies like~\cite{pmlr-v97-franc19a}, ~\cite{JMLR:v24:21-0048},~\cite{NIPS2011_4b6538a4}, and~\cite{Xia_2022_ACCV} have refined its methodologies and enhanced its efficacy, reinforcing its importance in research.

\subsection{Ad-hoc methods}
 Ad-hoc approaches~\citep{pmlr-v97-franc19a}~\citep{Xia_2022_ACCV} like SelectiveNet~\cite{geifman2019selectivenet} add an extra classification head for confidence prediction, enabling models to make selective predictions without changing the training process. Deep Gamblers~\citep{liu2019deep}, another ad-hoc method, introduces a unique loss function for predicting an ``uncertain" category, allowing models to manage uncertainty by ``gambling" on predictions. 

\paragraph{SelectiveNet}
SelectiveNet (SN)~\cite{geifman2019selectivenet} integrates the reject option directly into the architecture with a shared backbone and three heads: (i) a prediction head $f(x)$ for class probabilities or regression values, (ii) a selection head $g(x)$ that outputs a scalar confidence score in $[0,1]$ to decide whether to predict or abstain, and (iii) an auxiliary head $h(x)$, trained on the same task as $f(x)$ to expose the backbone to all samples and avoid overfitting to the confident subset. At inference, only $f(x)$ and $g(x)$ are used.

The selective loss function corresponding to Eq.~(\ref{eqn_opt}) is defined as:
\begin{equation}
\label{eqn_sn}
\mathcal{L}_{(f,g)} \triangleq R(f,g | S) + \lambda \ell(\overline{c},\mathbb{E}_P[g(x)]), 
\end{equation}
where $S$ is the labeled training dataset with $N$ samples, i.e., $S=\{(x_i, y_i)\}_{i=1}^N \subseteq (\mathcal{X}\times \mathcal{Y})^N$,
$\ell$ is a standard loss function, such as mean square loss, $\overline{c}$ is the target coverage, and $\lambda$ is a hyperparameter.

In parallel, the auxiliary head is trained with a standard supervised loss:
\begin{equation}
\mathcal{L}_{(h)} = \mathbb{E}_{P} [\ell(h(x), y)],
\end{equation}
where $h(x)\in\mathbb{R}^{B\times C}$ outputs class probabilities for batch size $B$ and $C$ classes.

The overall SelectiveNet objective is a convex combination of the selective and auxiliary losses:
\begin{equation}
\label{eqn_sn_loss}
\mathcal{L}_{SN} = \alpha \mathcal{L}_{(f,g)} + (1-\alpha)\mathcal{L}_{(h)},
\end{equation}
where $\alpha$ is a hyperparameter.


\paragraph{Deep Gamblers (DG)}
Deep Gamblers~\cite{liu2019deep} approach selective classification by reframing an $C$-class task as an $(C+1)$-class problem, where the additional class represents abstention. Inspired by portfolio theory, the model learns to allocate probability mass between betting on one of the $m$ classes and reserving some mass in the abstain class, analogous to a gambler splitting wealth between risky bets and a safe reserve.

Formally, let $f_\theta(x) \in \mathbb{R}^{C+1}$ denote the softmax output of the network. The Gambler’s loss is defined as:
\begin{equation}
\max_{\theta} \sum_{i=1}^B \log \Big( f_\theta(x_i){y_i} \cdot o + f_\theta(x_i)_{C+1} \Big),
\end{equation}
where $B$ is the batch size, $y_i$ the true label of $x_i$, and $o>0$ a hyperparameter called the odds. A higher $o$ encourages the network to be confident in inferring, and a low $o$ makes it less confident.

The abstention probability $g(x)=f_\theta(x)_{C+1}$ serves as the selection function. By tuning $o$ and calibrating thresholds on $g(x)$, the method achieves competitive selective prediction without modifying model architectures.

\subsection{Post-hoc methods}
Post-hoc methods like softmax response and MC-Dropout~\citep{geifman2017selective} assess prediction confidence without modifying the model. Softmax response evaluates confidence through models' probabilistic outputs, while MC-Dropout measures uncertainty by analyzing variance in multiple inference runs.

\paragraph{Softmax Response (SR)}~\cite{geifman2017selective} proposes to use the maximal activation in a classification network's softmax layer to define the confidence score, i.e., 
 \begin{equation}
 \label{eqn_sr}
     r(x) = \kappa(x)\triangleq \max_{i\in\mathcal{Y}}(p_i(x)), 
 \end{equation}
where $p(x)$ is a vector after the softmax layer, 
and its $i^{th}$ element  estimates the probability of output being the $i^{th}$ class.
This approach is simple and efficient, but not specifically optimized for selective prediction.
A recent finding from ~\cite{feng2022towards} also shows that SR can be a strong confidence indicator even when used combined with other selective prediction approaches such as SN and DG. 

~\cite{feng2022towards} introduced a novel framework wherein the model is trained using any of the ad-hoc methods like SN or DG, while evaluation of the model is done using the post-hoc methods like SR discarding the selection mechanism of ad-hoc methods. This hybrid framework has an edge over the standalone ad-hoc and post-hoc methods because the original selection mechanism of the ad-hoc method is suboptimal. 
However, unlike this framework which employs SR solely at inference time, our approach integrates SR directly into the training phase, thereby synchronizing the ad-hoc selection mechanism with the implicit uncertainty estimates from SR. This fundamental difference allows our method to improve both calibration and selective prediction performance while retaining the benefits of the ad-hoc selection head.

\section{Method}

We motivate the application of SR during inference, in conjunction with challenges currently faced by SN, leading to our proposed method SYNC loss which incorporates SR into training. We introduce how SYNC loss can be added to SN, and explain how it contributes to performance improvement. Additionally, we introduce an adjustable score function that can adapt to different datasets to further improve performance.

\subsection{Motivation}

\begin{figure}[!t]

\centering
\includegraphics[width=\linewidth]{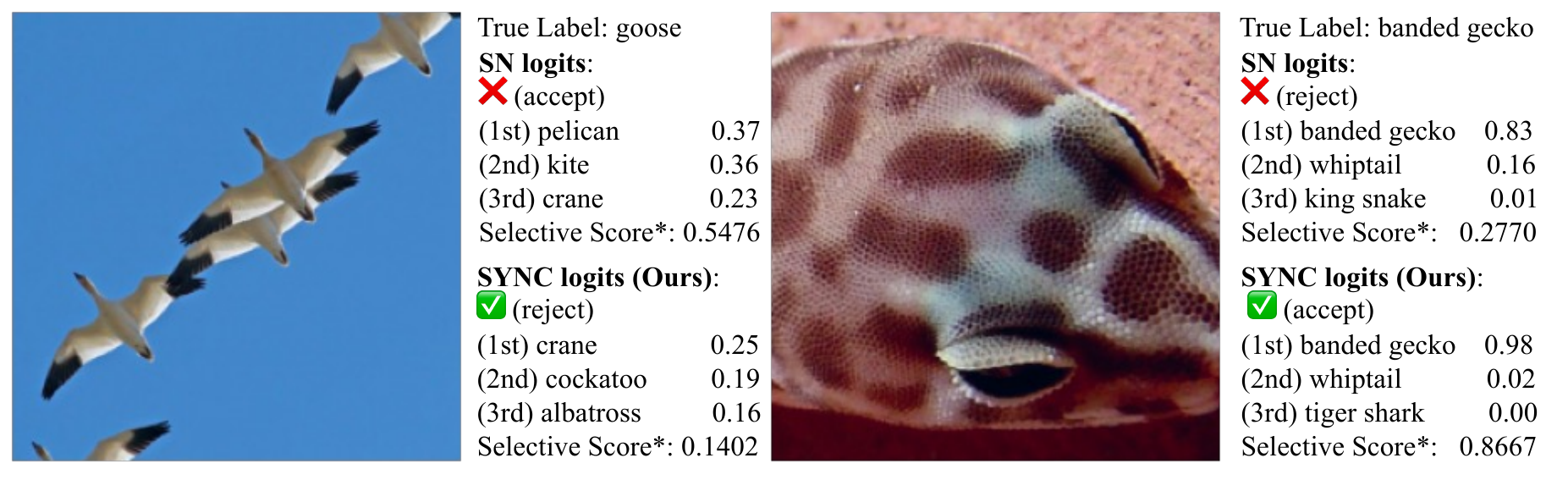}
\caption{\textbf{Examples of wrong rejection and wrong acceptance of SN from ImageNet100.} We use a threshold of 0.5. $^*$ implies that selective scores are normalized based on its ranking in the test set. Our approach calibrates the selective score by selective prior, therefore correctly avoid the wrong rejection and acceptance.}
\vspace{0.5cm}
\label{fig:eg1}

\end{figure}

Our motivation stems from an observed inconsistency between SN's selection mechanism and the SR. As illustrated in Fig. \ref{fig:eg1}, despite SN assigning a high selective score to a particular classification, SR indicates that the model exhibits low confidence in its prediction. Specifically, SR reveals that when multiple classes exhibit similar logits, the model experiences difficulty in discriminating between categories. This uncertainty manifests as a phenomenon we term \textit{selective prior}, wherein the model's decisions under such conditions demonstrate increased susceptibility to misclassification. These discrepancies indicate that relying on a single selective mechanism fails to adequately capture the model's true confidence levels.

A fundamental limitation contributing to these discrepancies lies in SN's loss function formulation. The selective loss function forces the network to maximize selective scores for training samples with lower training loss while minimizing scores for samples with higher training loss. However, during neural network training, models typically achieve substantially higher accuracy and lower loss on training sets compared to test sets \cite{zhang2021understanding}. For datasets with relatively low complexity, models can achieve training accuracies approaching 100\%.

Under these conditions, a selective loss function predicated on training loss becomes ineffective at identifying samples that present genuine classification challenges, consequently failing to assign appropriately low selective scores to such instances. This limitation manifests in our experimental observations (Section~\ref{sec:exp_sp}), where SN demonstrates degraded performance at low coverage levels. Therefore, enhancing SN's confidence estimation capabilities requires the integration of auxiliary information sources.

We propose incorporating the selective prior, traditionally utilized exclusively during inference, into SN's training phase. This integration is achieved through a novel loss function designed to minimize discrepancies between SR and selective scores. By leveraging SR as an additional information source during training, we enhance both the model's overall performance and its selective prediction capabilities.

\subsection{SYNC Loss}
\begin{figure}[!t]
\centering
\includegraphics[width=\linewidth]{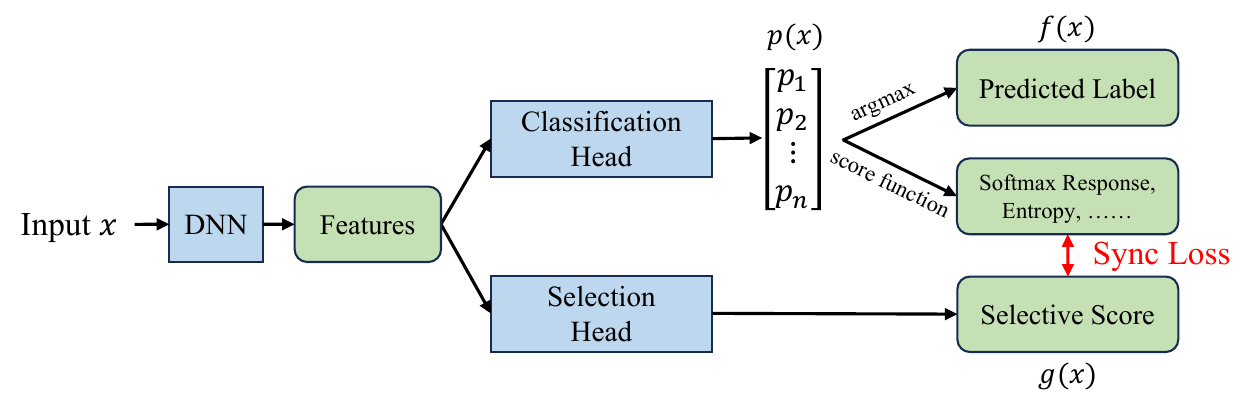}
\caption{Uncertainty-aware Selective Prediction Framework}
\vspace{0.5cm}
\label{fig:sn_pipeline}
\end{figure}

We use the SN framework to illustrate our idea. As depicted in Fig.~\ref{fig:sn_pipeline}, we incorporate the SR information by adding a new loss component, called SYNC loss, that aligns the selective score with SR-based uncertainty estimates. During the training process, we calibrate the estimated selective score using the SR output as a measure of prediction uncertainty. This approach enables the network to harness the additional generalization ability conferred by SR and addresses the overfitting issues inherent in SN on the training set.
This confidence calibration process is captured through a new loss function component defined as:
\begin{equation}
    \ell_{\text{sync}}(g(x), p(x)) \triangleq \ell(g(x), \text{score}(p(x))),    
\end{equation}
where $\ell$ is a standard loss function such as mean square loss (MSE); $g(x)$ is the selective model estimated selective score; $p(x)$ is the probability vector after the softmax layer;
and the $\text{score}(p(x))$ function is the estimation
of the uncertainty for the selective score $g(x)$ based on the
probability vector $p(x)$. 
While we use the SR function from Eq.
(\ref{eqn_sr}) for $\text{score}(p(x))$, this framework generalizes to other probabilistic estimation functions such as entropy function (detailed in Section
\ref{sec:score}).
Using the SN framework for selective prediction, our proposed loss function for optimization is:
\begin{equation}
  \mathcal{L}_{\text{SYNC}} \triangleq R(f,g|S) + \lambda \ell(\overline{c},\mathbb{E}_P[g(x)]) + \mu \ell_{\text{sync}}(g(x), p(x)),
\end{equation}
 where $\ell_\text{sync}$ is the proposed loss function for incorporating the SR information while training and $\mu$ is the balancing hyperparameter.
The overall loss becomes:
\begin{equation}
    \mathcal{L} = \alpha \mathcal{L}_{\text{SYNC}} + (1-\alpha) \mathcal{L}_{(h)},
\end{equation}
where $\mathcal{L}_{(h)}$ is the auxiliary loss from the SN framework and $\alpha$ is the hyperparameter.

The loss function, $\ell_\text{sync}$, optimizes the network's selection head output to align with the uncertainty estimates from the prediction head. This ensures that correctly classified samples have high selective scores when their predictions exhibit low uncertainty. During training, the selective head maximizes scores for samples that achieve both low training loss and low uncertainty from the score($\cdot$) function. This dual optimization yields a model that is well-calibrated in both confidence estimation and selective scoring.

\subsubsection{Effect of SYNC Loss}
SN suffers from misalignment between its selective scores and actual prediction uncertainty—it rejects samples with low prediction uncertainty while accepting those with high uncertainty. While \cite{feng2022towards} addressed this by discarding the selective score entirely in favor of SR, this creates a new problem: without confidence calibration, the model's softmax scores fail to reflect true correctness likelihood, leading to overconfident acceptance of samples that should be rejected. Our SYNC loss resolves both issues by synchronizing selective scores with uncertainty estimates.

The $\ell_{\text{sync}}$ loss calibrates selective scores using uncertainty estimates from the prediction head. This regularization term enhances both generalization and selective prediction capabilities of the underlying SN framework. Let $\nabla_f \mathcal{L}_{\text{SN}}$ and $\nabla_g \mathcal{L}_{\text{SN}}$
denote the gradients of the SN loss with respect to the prediction head $f$
 and selection head
$g$, respectively. The gradients of our proposed loss $\mathcal{L}_{\text{SYNC}}$ are:

\begin{equation}\label{eq1_}
    \nabla_f \mathcal{L}_{\text{SYNC}} = \nabla_f \mathcal{L}_{SN} + 2 * \nabla (p(x))[p(x) - g(x)],
\end{equation}
and,
\begin{equation}\label{eq2_}
    \nabla_g \mathcal{L}_{\text{SYNC}} = \nabla_g \mathcal{L}_{SN} + 2 * [g(x) - p(x)],
\end{equation}
where we use MSE for $\ell_{\text{sync}}$ and maximum softmax probability for
$\text{score}(p(x))$ for clarity. The regularization term $g(x) - p(x)$ in Eq.~\ref{eq2_} aligns the selection head
$g$ with the model's uncertainty estimates, directly improving selective prediction performance and addressing the misalignment shown in Fig.~\ref{fig:eg1}. The regularization term $\nabla (p(x))[p(x) - g(x)]$ in Eq.~\ref{eq1_} calibrates the prediction head $f$, enhancing the model's overall generalization capability.

\subsection{Score Function}
\label{sec:score}
The score function is a pivotal component in our model, tasked with estimating uncertainty from the probabilistic outputs of the network. Traditional methods often rely on SR for this purpose, but our approach extends beyond this convention, exploring a range of alternative functions such as negative entropy ~\cite{negative_entropy}, spike entropy~\citep{kirchenbauer2023watermark}, and even sigmoid and exponential functions. We introduce a score function called Softmax Power(SMP) Score. It is mathematically represented as:
\begin{equation}
\text{score}(p(x)) = {\left(max(p(x))\right)}^\gamma,
\end{equation}
where $p(x)$ denotes the softmax probability vector generated from a set of inputs, and $\gamma$ represents a user-defined hyperparameter controlling the power transformation. The SMP Score offers flexibility in adjusting the degree of emphasis placed on the maximum probability, allowing us to fine-tune the scoring process according to specific application requirements. 
Collectively, these scoring functions—including SR, negative entropy, spike entropy, sigmoid/exponential transforms, and our SMP—provide a more nuanced way to interpret probability distributions, thereby providing a richer understanding of the model's confidence in its predictions. 

\subsection{Lipschitz \& Smoothness Analysis}
\paragraph{Notation.}
Let $\mathcal X\subset\mathbb R^{d}$ be the input space and
$\mathcal Y=\{1,\dots,C\}$ the label set with $C\ge2$ classes.
For any input $x\in\mathcal X$ the backbone network produces raw
\emph{logits} as
\[
        z(x)\;=\;\bigl(z_1(x),\dots,z_C(x)\bigr)\in\mathbb R^{C},
\]
which are converted to class–posterior probabilities via the
softmax.
\[
        p(x)\;=\;\operatorname{softmax}\!\bigl(z(x)\bigr)
              \;=\;
              \bigl(p_1(x),\dots,p_C(x)\bigr)
              \;\in\;
              \Delta^{\,C-1}.
\]

\vspace{2pt}
\noindent\textbf{Probability simplex.}
Throughout we use
\begin{align}
\Delta^{\,C-1}
\;=\;
\bigl\{
        u\in[0,1]^{C}\;
        \bigm|\;
        \textstyle\sum_{i=1}^{C} u_i = 1
\bigr\}
\label{eq:simplex}
\end{align}
to denote the closed $(C\!-\!1)$-dimensional probability simplex
embedded in $\mathbb R^{C}$.

\paragraph{Selector head.}
A separate selector network
$g_{\theta}:\mathbb R^{d}\!\to[0,1]$,
parameterised by $\theta$, decides whether to \emph{accept} the
backbone prediction ($g_\theta(x)\approx1$) or \emph{reject} it
($g_\theta(x)\approx0$).

\paragraph{SYNC loss.}
The training objective is:
\[
\mathcal L_{\text{SYNC}}(x)
\;=\;
\mathcal L_{\text{SN}}(x)\;+\;
\mu\;\bigl\|g_{\theta}(x)\;-\;s_{\gamma}\!\bigl(p(x)\bigr)\bigr\|_2^2,
\]
where $\mathcal L_{\text{SN}}$ is the original SelectiveNet loss,
$\mu>0$ is a trade-off weight, and
\[
s_{\gamma}\!\bigl(p(x)\bigr)
\;=\;
\Bigl(\max_{i}\,p_i(x)\Bigr)^{\gamma},
\qquad
\gamma>0,
\]
is the \emph{Softmax-Power (SMP) score} introduced in
Section 3.3.

\paragraph{Bounded-logit assumption.}
Following common practice in margin-based analyses, we assume that
for all $x\in\mathcal X$
\begin{equation}
        \|z(x)\|_{\infty}\;\le\;B,
        \qquad\text{for some }B>0.
        \label{eq:logit_bound}
\end{equation}
This can be enforced with weight decay, spectral normalisation, or
explicit logit clipping.

\subsubsection{Lipschitz / Hölder continuity of the SMP score}
\label{sec:lipschitz_smp}
The \emph{Softmax–Power (SMP)} score is
\begin{equation}
s_\gamma(u)=\bigl(\max_{i} u_i\bigr)^{\gamma},
\qquad
u\in\Delta^{\,C-1},\;
\gamma>0.
\end{equation}
where $\Delta^{\,C-1}$ is the closed probability simplex
defined in~\eqref{eq:simplex}.
The following lemma quantifies how sharply $s_\gamma$ can change. Let $m(u)=\max_i u_i$. Note that $m(u)\in[1/C,1]$ for all $u\in\Delta^{\,C-1}$. For a matrix $A$, $\lambda_{\max}(A)$ denotes its spectral norm (largest singular value). 
For symmetric $A$, this equals its largest eigenvalue.
\begin{lemma}[Lipschitz continuity of $s_\gamma$ on the simplex]
\label{lem:smp_continuity}
For $u\in\Delta^{C-1}$ and $\gamma>0$, define $s_\gamma(u):=\bigl(\max_i u_i\bigr)^\gamma$.
Then $s_\gamma$ is $L_\gamma$–Lipschitz on $\Delta^{C-1}$ in the $\ell_\infty$ norm, with
\[
L_\gamma \;=\;
\begin{cases}
\gamma, & \gamma\ge 1,\\[2pt]
\gamma\,C^{\,1-\gamma}, & 0<\gamma<1.
\end{cases}
\]
Equivalently, for all $u,v\in\Delta^{C-1}$,
$\;|s_\gamma(u)-s_\gamma(v)| \le L_\gamma \,\|u-v\|_\infty$.
\end{lemma}

\paragraph{Interpretation and role.}
Lemma~\ref{lem:smp_continuity} guarantees that the SMP target $s_\gamma\!\bigl(p(x)\bigr)$ varies smoothly with the softmax vector $p(x)$:
small perturbations in $p(x)$ induce at most $L_\gamma$-scaled changes in the target.
Consequently, the SYNC regulariser $\,\mu\!\left(g_\theta(x)-s_\gamma\!\bigl(p(x)\bigr)\right)^2$ does not overreact to minor stochastic fluctuations in predicted probabilities, helping to avoid gradient spikes and promoting more stable optimization.

\paragraph{Intuition.}
Let $f(t)=t^\gamma$ be the scalar power map on $[1/C,1]$.
Since $s_\gamma(u)=f(m(u))$ depends only on the largest entry $m(u):=\max_i u_i$,
changing $u$ to $v$ changes this scalar from $a:=m(u)$ to $b:=m(v)$, and $|a-b|\le\|u-v\|_\infty$.
The sensitivity of $f$ at a scalar $t\in[1/C,1]$ is $|f'(t)|=\gamma t^{\gamma-1}$.
Hence the worst-case slope on $[1/C,1]$ is
$\gamma$ if $\gamma\ge 1$ and $\gamma C^{\,1-\gamma}$ if $0<\gamma<1$,
which, combined with $|a-b|\le\|u-v\|_\infty$, yields the stated bounds.

\paragraph{Proof.}
Refer to Supplementary Material~\S I-A.

\begin{corollary}[Sharpness--stability trade-off for $\gamma$]\label{cor:sharp-stability}
Let $s_\gamma$ be as in Lemma~\ref{lem:smp_continuity}, with Lipschitz modulus
\[
L_\gamma=\begin{cases}
\gamma, & \gamma\ge 1,\\
\gamma\,C^{\,1-\gamma}, & 0<\gamma<1.
\end{cases}
\]
Consider the composite map $\mathcal F_\gamma(x):=s_\gamma\!\bigl(\mathrm{softmax}(z(x))\bigr)$. 
If
\begin{equation}\label{eq:contractivity-cond}
\|\nabla_x z(x)\|_2\;\lambda_{\max}\!\bigl(\nabla_z\,\mathrm{softmax}(z)\bigr)\;L_\gamma \;\le\; 1
\quad\text{for all $x$,}
\end{equation}
then $\mathcal F_\gamma$ is nonexpansive (i.e., $1$-Lipschitz). In particular, if
$\lambda_{\max}\!\bigl(\nabla_z\,\mathrm{softmax}(z)\bigr)\le 1/B$, it suffices to choose $\gamma$ so that
\begin{equation}\label{eq:gamma_bound_piecewise}
\boxed{
\begin{aligned}
\gamma \;&\le\; \dfrac{B}{\|\nabla_{x} z(x)\|_{2}}, && \text{if }\gamma\ge 1,\\[6pt]
\gamma\,C^{\,1-\gamma} \;&\le\; \dfrac{B}{\|\nabla_{x} z(x)\|_{2}}, && \text{if }0<\gamma<1.
\end{aligned}}
\end{equation}
\end{corollary}
\paragraph{Proof.}
Refer to Supplementary Material~\S I-B.

\subsection{Global Smoothness of the \texorpdfstring{$\mathcal{L}_{\mathrm{SYNC}}$}{L\_SYNC} Objective}
\label{sec:prop-smooth-full}

\textbf{Standing hypotheses.}
\begin{enumerate}[label=(H\arabic*),leftmargin=30pt]
    \item \label{hyp:smooth-sn}
          (\emph{Smooth baseline}) \;
          $\|\nabla_{\theta}^{2}\mathcal L_{\mathrm{SN}}(x)\|_{2}\le L$
          for some constant $L<\infty$ and every $x\in\mathcal X$.
    \item \label{hyp:bound-jac}
          (\emph{Bounded selector Jacobian}) \;
          $\|\nabla_{\theta}g_\theta(x)\|_{F}\le G_{\!*}$
          for all $x\in\mathcal X$.
\end{enumerate}

\begin{proposition}
\label{prop:smooth-detailed}
Under hypotheses \ref{hyp:smooth-sn}--\ref{hyp:bound-jac},
the Hessian with respect to~$\theta$ satisfies
\begin{equation}
\nabla_{\theta}^{2}\mathcal L_{\mathrm{SYNC}}(\theta;x)
  \;\preccurlyeq\;
  \bigl(L + 2\mu G_{\!*}^{2}\bigr)\,I_{d},
  \qquad\forall\bigl(\theta,x\bigr)\in\Theta\times\mathcal X,
  \label{eq:hessian-bound}
\end{equation}
where $I_{d}$ is the $d\times d$ identity matrix.
Consequently, the gradient mapping
$\theta\mapsto\nabla_{\theta}\mathcal L_{\mathrm{SYNC}}(\theta;x)$
is $(L+2\mu G_{\!*}^{2})$–Lipschitz, i.e.
\[
\bigl\|\nabla_{\theta}\mathcal L_{\mathrm{SYNC}}(\theta_1;x)
      -\nabla_{\theta}\mathcal L_{\mathrm{SYNC}}(\theta_2;x)\bigr\|_{2}
  \;\le\;
  \bigl(L+2\mu G_{\!*}^{2}\bigr)\,
  \|\theta_1-\theta_2\|_{2}.
\]
\end{proposition}

Therefore, adding the quadratic SYNC regulariser does not break smoothness; it simply enlarges the global constant from $L$ to $(L+2\mu G_{\!*}^{2})$, after which standard optimisation guarantees continue to apply unchanged.
\paragraph{Proof.}
Refer to Supplementary Material~\S I-C.

\section{Experiments}
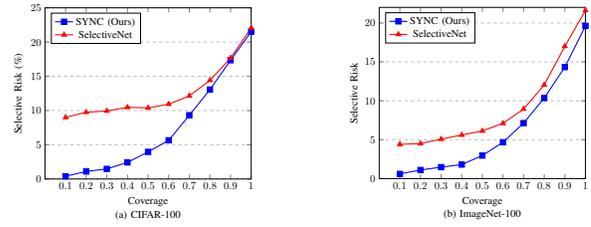
\begin{figure}[!t]
    \centering
    \begin{minipage}[t]{0.49\linewidth}
        \centering
        \begin{tikzpicture}[scale=0.40]
\begin{axis}[
    title={(a) CIFAR-100},
    title style={align=center, at={(axis description cs:0.5,-0.2)}, anchor=north},
    xlabel={Coverage},
    ylabel={Selective Risk (\%)},
    xmin=0, xmax=1,
    ymin=0, ymax=25,
    xtick={0.1,0.2,0.3,0.4,0.5,0.6,0.7,0.8,0.9,1},
    ytick={0,5,10,15,20,25},
    legend pos=north west,
    ymajorgrids=true,
    grid style=dashed,
]

\addplot[
    color=blue,
    mark=square*,
    mark size=2.5pt,
    line width=1pt,
    ]
    coordinates {
    (0.1,0.4)(0.2,1.1)(0.3,1.47)(0.4,2.42)(0.5,3.96)(0.6,5.65)(0.7,9.31)(0.8,13.06)(0.9,17.34)(1,21.51)
    };
    \addlegendentry{SYNC (Ours)}

\addplot[
    color=red,
    mark=triangle*,
    mark size=2.5pt,
    line width=1pt,
    ]
    coordinates {
    (0.1,9)(0.2,9.75)(0.3,9.93)(0.4,10.47)(0.5,10.38)(0.6,10.93)(0.7,12.14)(0.8,14.41)(0.9,17.66)(1,22.06)
    };
    \addlegendentry{SelectiveNet}


\end{axis}
\end{tikzpicture}
    \end{minipage}%
    \hfill
    \begin{minipage}[t]{0.49\linewidth}
        \centering
        \begin{tikzpicture}[scale=0.40]
\begin{axis}[
    title={(b) ImageNet-100},
    title style={align=center, at={(axis description cs:0.5,-0.2)}, anchor=north},
    xlabel={Coverage},
    ylabel={Selective Risk},
    xmin=0, xmax=1,
    ymin=0, ymax=22,
    xtick={0.1,0.2,0.3,0.4,0.5,0.6,0.7,0.8,0.9,1},
    ytick={0,5,10,15,20},
    legend pos=north west,
    ymajorgrids=true,
    grid style=dashed,
]

\addplot[
    color=blue,
    mark=square*,
    mark size=2.5pt,
    line width=1pt,
    ]
    coordinates {
    (0.1,0.6)(0.2,1.1)(0.3,1.47)(0.4,1.8)(0.5,2.96)(0.6,4.67)(0.7,7.13)(0.8,10.35)(0.9,14.33)(1,19.62)
    };
    \addlegendentry{SYNC (Ours)}

\addplot[
        color=red,
    mark=triangle*,
    mark size=2.5pt,
    line width=1pt,
    ]
    coordinates {
    (0.1,4.4)(0.2,4.5)(0.3,5.07)(0.4,5.6)(0.5,6.12)(0.6,7.1)(0.7,8.94)(0.8,12.03)(0.9,17)(1,21.62)
    };
    \addlegendentry{SelectiveNet}

\end{axis}
\end{tikzpicture} 
    \end{minipage}

\caption{Selective risk coverage (lower the better) curves comparing SN and SYNC on (a) CIFAR-100 and (b) ImageNet-100.}
\vspace{0.5cm}
\label{fig:score}
\end{figure}

\begin{table}
\caption{Comparison of the selective classification error at 100\% coverage showing the generalizability of our approach over CIFAR-100 and ImageNet100 datasets.}
\vspace{0.3cm}
\label{tab:cov1}
\centering
\begin{tabular}{@{}ccc@{}}
\toprule
Model        & ImageNet-100            & CIFAR-100               \\ \midrule
Vanilla      & 21.80$\pm$0.22          & 22.27$\pm$0.18          \\
SelectiveNet & 21.07$\pm$0.67          & 21.68$\pm$0.22          \\
Deep Gambler & 20.78$\pm$0.16          & 21.70$\pm$0.38           \\
SYNC$^{\gamma=1}$(Ours)  & \textbf{20.19$\pm$0.49} & \textbf{21.22$\pm$0.21} \\ \bottomrule
\end{tabular}

\end{table}



\begin{table*}[htb]
\caption{Comparison of the selective prediction error between SN, DG and SYNC
with the original selection mechanisms vs. using SR on CIFAR-100. \textbf{Bold} data is used to indicate that this represents the best result among several methods using the same selective mechanism. \underline{Underlined} data signifies that it is the best considering different selective mechanisms. All values in the table are expressed as mean \(\pm\) standard deviation.
}
\label{tab:cov_cifar100}
\centering
\resizebox{0.9\linewidth}{!}{

\begin{tabular}{@{}ccccccc@{}}
\toprule
\multirow{2}{*}{Cov.}  & \multicolumn{3}{c}{\textbf{w/o SR}} & \multicolumn{3}{c}{\textbf{w/ SR}} \\
\cmidrule(r){2-4} \cmidrule(l){5-7} & \textbf{SN} & \textbf{DG}& \textbf{SYNC$^{\gamma=0.5}$ (Ours)} & \textbf{SN} & \textbf{DG} & \textbf{SYNC$^{\gamma=2.5}$ (Ours)} \\
\midrule
100  & 22.06$\pm$0.15 & 22.39$\pm$0.33 & \textbf{21.51$\pm$0.27} & 22.06$\pm$0.15 & 22.39$\pm$0.33 & \underline{\textbf{21.36$\pm$0.21}} \\
90   & 17.66$\pm$0.35 & 19.33$\pm$0.32 & \textbf{17.34$\pm$0.35} & 16.40$\pm$0.12 & 17.03$\pm$0.33 & \underline{\textbf{15.97$\pm$0.17}} \\
80   & 14.41$\pm$0.40 & 17.40$\pm$0.32 & \textbf{13.06$\pm$0.35} & 12.25$\pm$0.25 & 12.49$\pm$0.19 & \underline{\textbf{11.73$\pm$0.14}} \\
70   & 12.14$\pm$0.58 & 16.06$\pm$0.26 & \textbf{ 9.31$\pm$0.09} &  8.90$\pm$0.27 &  9.03$\pm$0.07 & \underline{\textbf{ 8.13$\pm$0.19}} \\
60   & 10.93$\pm$0.40 & 14.82$\pm$0.26 & \textbf{ 5.65$\pm$0.26} &  6.43$\pm$0.01 &  6.57$\pm$0.35 & \underline{\textbf{ 5.38$\pm$0.22}} \\
50   & 10.38$\pm$0.35 & 13.54$\pm$0.34 & \textbf{ 3.96$\pm$0.29} &  4.18$\pm$0.10 &  4.40$\pm$0.43 & \underline{\textbf{ 3.52$\pm$0.09}} \\
40   & 10.47$\pm$0.31 & 12.73$\pm$0.34 & \textbf{ 2.42$\pm$0.56} &  2.67$\pm$0.06 &  2.70$\pm$0.30 & \underline{\textbf{ 2.20$\pm$0.10}} \\
30   &  9.93$\pm$0.41 & 11.30$\pm$0.17 & \textbf{ 1.47$\pm$0.63} &  2.00$\pm$0.15 &  1.80$\pm$0.09 & \underline{\textbf{ 1.37$\pm$0.03}} \\
20   &  9.75$\pm$0.06 & 10.00$\pm$0.50 & \textbf{ 1.10$\pm$0.51} &  1.15$\pm$0.29 &  1.45$\pm$0.31 & \underline{\textbf{ 1.00$\pm$0.14}} \\
10   &  9.00$\pm$0.53 &  9.50$\pm$0.57 & \underline{\textbf{ 0.40$\pm$0.56}} &  1.10$\pm$0.15  &  1.10$\pm$0.12 & \textbf{ 0.70$\pm$0.12} \\
\bottomrule
\end{tabular}
}
\vspace{0.2cm}

\end{table*}

\begin{table*}[htb]
\caption{Comparison of the selective prediction error between SN, DG and SYNC with the original selection mechanisms vs. using SR on ImageNet-100.
}
\label{tab:cov_imagenet100}
\centering
\resizebox{0.9\linewidth}{!}{
\begin{tabular}{@{}ccccccc@{}}
\toprule
\multirow{2}{*}{Cov.} & \multicolumn{3}{c}{\textbf{w/o SR}} & \multicolumn{3}{c}{\textbf{w/ SR}} \\
\cmidrule(r){2-4} \cmidrule(l){5-7} & \textbf{SN} & \textbf{DG}  & \textbf{SYNC$^{\gamma=1}$ (Ours)} & \textbf{SN} & \textbf{DG} &  \textbf{SYNC$^{\gamma=1}$ (Ours)} \\
\midrule
100  & 21.07$\pm$0.67 & 20.78$\pm$0.16 & \underline{\textbf{20.19$\pm$0.49}} & 21.07$\pm$0.67 &20.78$\pm$0.16 & \underline{\textbf{20.19$\pm$0.49}} \\
90   & 16.27$\pm$0.68 & 19.33$\pm$0.09 & \underline{\textbf{14.79$\pm$0.54}} & 15.69$\pm$0.66 &15.41$\pm$0.12 & \textbf{15.17$\pm$0.36} \\
80   & 11.42$\pm$0.53 & 19.08$\pm$0.23 & \underline{\textbf{10.61$\pm$0.29}} & \textbf{11.25$\pm$0.58} &11.81$\pm$0.32 & 11.41$\pm$0.29 \\
70   & 8.27$\pm$0.59  & 18.92$\pm$0.21 & \underline{\textbf{7.33$\pm$0.26 }} & \textbf{8.45$\pm$0.38}  &9.57$\pm$0.33  & 8.95$\pm$0.54 \\
60   & 6.41$\pm$0.60  & 18.94$\pm$0.21 & \underline{\textbf{4.96$\pm$0.41 }} & \textbf{6.28$\pm$0.29}  &7.80$\pm$0.15  & 7.18$\pm$0.40 \\
50   & 5.51$\pm$0.58  & 18.85$\pm$0.27 & \underline{\textbf{3.25$\pm$0.38 }} & \textbf{4.51$\pm$0.02}  &6.17$\pm$0.21  & 5.60$\pm$0.32 \\
40   & 5.02$\pm$0.51  & 18.90$\pm$0.63 & \underline{\textbf{2.08$\pm$0.26 }} & \textbf{3.08$\pm$0.10}  &4.58$\pm$0.33  & 4.23$\pm$0.18 \\
30   & 4.62$\pm$0.41  & 18.84$\pm$0.34 & \underline{\textbf{1.69$\pm$0.28 }} & \textbf{2.13$\pm$0.29}  &3.53$\pm$0.24 & 3.24$\pm$0.30 \\
20   & 4.37$\pm$0.15  & 18.73$\pm$0.49 & \textbf{1.27$\pm$0.15 } & \underline{\textbf{0.93$\pm$0.06}}  &2.67$\pm$0.21 & 2.33$\pm$0.12 \\
10   & 4.00$\pm$0.53     & 17.53$\pm$0.31 & \textbf{0.93$\pm$0.31 } & \underline{\textbf{0.20$\pm$0.20}}  &2.00$\pm$0.60 & 0.80$\pm$0.40 \\
\bottomrule
\end{tabular}
}

\vspace{0.2cm}
\end{table*}

This section assesses the effectiveness of our proposed approach through a series of experiments.
Through our experiments, we show that: (1) Our proposed SYNC loss, compared to SN, achieves significant improvement by using softmax response to calibrate selective scores; (2) By leveraging selective prior, our method enhances the model's generalization capabilities; (3) Our method achieves improvements over existing works across different datasets and under various selective mechanisms.

We begin by comparing the selective prediction of SYNC against SN. We then showcase the efficacy of our approach in enhancing the generalizability of the model regardless of the selective prediction i.e. at 100\% coverage. We conduct thorough comparisons with existing methodologies in selective classification, demonstrating that our proposed technique is highly competitive against state-of-the-art (SOTA) methods. In the end, we perform a comparative analysis with the existing approaches, demonstrating that our proposed method outperforms the existing baselines. Lastly, we conduct an ablation study to verify the rationality of our choice of the hyper-parameters. We perform our experiments on CIFAR-100, Stanford Cars, and ImageNet-100 datasets. In the experiments ``w/o SR" implies that the original selective mechanism of the method is used and ``w/ SR" implies the method suggested in ~\cite{feng2022towards} where we discard the methods' original selective mechanism with SR. Dataset information and experiment details are in the supplementary material \S II-A and II-B, respectively.
\begin{table*}[!htb]
\caption{Comparison of the selective prediction error between SN, DG and SYNC with the original selection mechanisms vs. using SR on Stanford Cars.
}
\label{tab:cov_cars}
\centering
\resizebox{0.9\linewidth}{!}{
\begin{tabular}{@{}ccccccc@{}}
\toprule
\multirow{2}{*}{Cov.} & \multicolumn{3}{c}{\textbf{w/o SR}} & \multicolumn{3}{c}{\textbf{w/ SR}} \\
\cmidrule(r){2-4} \cmidrule(l){5-7} & \textbf{SN} & \textbf{DG}  & \textbf{SYNC$^{\gamma=1}$ (Ours)} & \textbf{SN} & \textbf{DG} &  \textbf{SYNC$^{\gamma=1}$ (Ours)} \\
\midrule
100  & 40.01$\pm$0.36 & 39.35$\pm$0.07 &  \underline{\textbf{39.11$\pm$0.26}} & 40.01$\pm$0.36 &39.35$\pm$0.07 & \underline{\textbf{39.11$\pm$0.26}} \\
90   & 34.28$\pm$0.24 & 33.44$\pm$0.05 &  \textbf{33.22$\pm$0.22} & 34.16$\pm$0.39 &34.22$\pm$0.23 & \underline{\textbf{33.18$\pm$0.36}} \\
80   & 28.43$\pm$0.15 & 29.15$\pm$0.16 &  \underline{\textbf{27.71$\pm$0.09}} & 28.35$\pm$0.37 &28.81$\pm$0.36 & \textbf{27.72$\pm$0.50} \\
70   & 23.71$\pm$0.39 & 26.15$\pm$0.40 &  \textbf{22.72$\pm$0.33} & 23.08$\pm$0.35 &\underline{\textbf{23.00$\pm$0.31}} & 23.05$\pm$0.83 \\
60   & 19.34$\pm$0.65 & 24.2 $\pm$0.32 &  \textbf{18.75$\pm$0.41} & 18.92$\pm$0.94 &\underline{\textbf{17.72$\pm$0.04}} & 18.81$\pm$0.83 \\
50   & 15.53$\pm$0.84 & 21.76$\pm$0.38 &  \textbf{15.26$\pm$0.71} & 15.43$\pm$1.03 &\underline{\textbf{14.42$\pm$0.21}} & 15.17$\pm$0.95 \\
40   & 12.47$\pm$0.77 & 20.36$\pm$0.49 &  \textbf{12.45$\pm$0.83} & 13.01$\pm$0.91 &12.59$\pm$0.07 & \underline{\textbf{12.42$\pm$0.71}} \\
30   & 10.42$\pm$0.60 & 18.73$\pm$0.54 &  \textbf{10.22$\pm$0.71} & 10.55$\pm$0.81 &10.61$\pm$0.07 & \underline{\textbf{\textbf{10.07$\pm$0.67}}} \\
20   &  7.96$\pm$0.53 & 18.1 $\pm$0.83 &  \textbf{ 7.93$\pm$0.60} &  8.49$\pm$0.77 & 8.21$\pm$0.04 & \underline{\textbf{ 7.90$\pm$0.48}} \\
10   &  6.03$\pm$0.58 & 15.55$\pm$0.57 &  \underline{\textbf{ 5.35$\pm$0.44}} &  6.03$\pm$0.40 & 7.34$\pm$0.09 & \textbf{ 5.72$\pm$0.26} \\
\bottomrule
\end{tabular}
}
\vspace{0.2cm}

\end{table*}

\begin{table}[!hbt] 
\caption{Analyzing different scoring functions, results with and without SR}
\label{Scoring_fn}
\centering
\begin{tabular}{lllll}
\toprule
\multirow{2}{*}{\shortstack{\textbf{Scoring function/}\\\textbf{Coverage}}} & \multicolumn{2}{c}{\textbf{Negentropy$^{\alpha=1.0}$}} & \multicolumn{2}{c}{\textbf{Max$^{\alpha=1.0}$}} \\ 
\cmidrule(r){2-3} \cmidrule(l){4-5}                                 & \textbf{Without SR} & \textbf{With SR} & \textbf{Without SR} & \textbf{With SR} \\ \midrule
1.0                                & 21.87               & 21.87            & \textbf{21.22}      & \textbf{21.22}   \\
0.9                                & 17.94               & 16.44            & \textbf{16.54}      & \textbf{15.82}   \\
0.8                                & 15.16               & 12.17            & \textbf{11.98}      & \textbf{11.74}   \\
0.7                                & 12.83               & 8.9              & \textbf{10.00}      & \textbf{8.6}     \\
0.6                                & 10.53               & 6.5              & \textbf{8.90}       & \textbf{5.95}    \\ \bottomrule
\end{tabular}
\vspace{0.3cm}
\end{table}


\begin{table}[]
\caption{Analyzing different values of SMP hyper-parameter $\gamma$ for "without SR" and "with SR" mechanisms.}
\label{SMP-a}
\centering
\begin{tabular}{lcccccc}
\toprule
\multirow{2}{*}{\textbf{Coverage/SMP($\gamma$)}} & \multicolumn{3}{c}{\textbf{Without SR}} & \multicolumn{3}{c}{\textbf{With SR}} \\ 
\cmidrule(r){2-4} \cmidrule(l){5-7}  & 0.5   & 1.0   & 2.5   & 0.5   & 1.0   & 2.5   \\ \midrule
1.0                      & 21.51 & \textbf{21.22} & 21.36 & 21.51 & \textbf{21.22} & 21.36 \\
0.9                      & 17.34 & \textbf{16.54} & 18.02 & 16.69 & \textbf{15.82} & 15.97 \\
0.8                      & 13.06 & \textbf{11.98} & 15.91 & 12.41 & 11.74 & \textbf{11.73} \\
0.7                      & \textbf{9.31}  & 10.00 & 14.04 & 8.99  & 8.60  & \textbf{8.13}  \\
0.6                      & \textbf{6.22}  & 8.90  & 12.83 & 6.45  & 5.95  & \textbf{5.38}  \\
\bottomrule
\end{tabular}
\vspace{0.1cm}

\end{table}


\subsection{Results}

\subsubsection{Softmax Response Guided Selective Score}
We initiated our experimentation by comparing the efficacy of our approach against the original SN framework. The risk-coverage curve (refer Fig. \ref{fig:score}) shows that SN tends to exhibit suboptimal performance at low coverage levels. This is due to the optimization process of SN that ignores the samples that hard to classify but have a relatively small training loss. In contrast, guided by the selective prior (Softmax Response), our method significantly improves the selective prediction performance on all coverages, especially on low coverages.
Our model exhibits greater robustness in the view of uncertainty and variability, providing better selection prediction performance. In Table~\ref{tab:cov1}, we compare the test risk of the proposed method with existing methods SN and DG on CIFAR-100, Stanford Cars and ImageNet-100 datasets. We observe that our method has achieved better results than each of these methods on all the datasets by a margin.

\subsubsection{Selective Prediction}
\label{sec:exp_sp}

We compare the selective prediction performance of our method and other approaches across four datasets. For each of the methods, we evaluate it selective accuracy with the original selection mechanisms vs. using Softmax Response, i.e. , for SN and SYNC, we use the selective score, and for DG, we use the abstention logit. For each of the datasets, we report the selective accuracy under different test coverages from 10\% to 100\%.

\textbf{CIFAR-100} As shown in Table~\ref{tab:cov_cifar100}, we first note that our method, when employing the original selective mechanism, has significantly improved the accuracy across various test coverages compared to the original SN.
As claimed in \cite{feng2022towards} both SR and SN achieve better performance across various test coverages when SR serves as a selective mechanism than their original selective mechanism. This is similarly applied to other datasets.
Note that, our method, even without utilizing SR, has achieved results comparable to SN when it employs SR. As we claimed before, our method can significantly improve the model's generalizability, when our method is also equipped with SR.
As we claimed before, our method significantly improves the model's generalizability. When our method is also equipped with SR, we achieve even better results, except at the 10\% and 20\% test coverages. 
Interestingly, without SR, DG achieved much poorer results compared to other methods. By observing the training curves, we found that DG with the original selective mechanism tends to overfit at lower test coverages, leading to a significant decline in performance.

\textbf{ImageNet-100} To verify the scalability of our method, we further conducted our experiments on ImageNet-100 as shown in Table~\ref{tab:cov_imagenet100}. We observed that, unlike on CIFAR-100, our method, without utilizing SR and relying solely on the original selective mechanism, achieved much better results compared to SN and DG using SR. This further validates our claim. By incorporating prior knowledge about SR during the training process, we eliminate the need to use it during inference. Additionally, introducing SR in the training phase enables us to learn a more generalized model, ultimately leading to better results. Regarding the difference in results compared to CIFAR-100, we believe that since ImageNet is more comprehensive, it results in a more calibrated Softmax response during the training phase, which in turn leads to better model training.

\textbf{Stanford Cars} To verify the adaptability and robustness of our method across diverse and challenging domains, we extend our experiments on Stanford Cars dataset as shown in Table~\ref{tab:cov_cars}. We observe a similar trend as for ImageNet-100, where our method, employing the original selective mechanism, consistently achieves results closely aligned with its SR counterpart. It also attains SOTA results at 10\%, 80\%, and 100\% coverages. and achieves results close to methods with SR. Notably, our method, after employing SR, achieved SOTA results in the majority of coverage metrics. This reaffirms the effectiveness and versatility of our proposed approach across varied datasets and selective mechanisms.

For \textbf{convergence analysis}, refer to supplementary material \S II-C. For additional examples, refer to supplementary material \S II-D.
\subsection{Ablation Studies}\label{ablation_}

In this section, we conduct a series of ablation studies to validate the effectiveness of the proposed Softmax Power (SMP) scoring function, denoted as \( p(x) \), and investigate the impact of the hyper-parameter \( \gamma \) in the SMP scoring function. We compare our method against negative entropy scoring function on the CIFAR-100 dataset.
\subsubsection{Scoring Functions}
We first compare the SMP scoring function with the negative entropy scoring function. For a comprehensive analysis, we evaluate the performance under various selection mechanisms, specifically examining scenarios with and without the SR mechanism. The detailed experimental result is shown in Table \ref{Scoring_fn} ($\alpha$ is the weight balancing hyper-parameter). 
The findings consistently demonstrate that SMP outperforms negative entropy regardless of the selection mechanism (with or without SR). This enhancement in performance underscores the efficacy of the SMP function in managing model uncertainties more effectively than traditional methods.
\subsubsection{Impact of Hyper-parameter $\gamma$ on SMP Performance}
To further explore the effectiveness of the SMP scoring function, we vary the hyper-parameter \( \gamma \), which plays a critical role in adjusting the sensitivity of the scoring function to prediction confidence levels. 
We conducted experiments with different values of \( \gamma \) and recorded the performance variations on the CIFAR-100 dataset. The outcomes of these experiments are detailed in Table \ref{SMP-a}. The results indicate that the optimal value of \( \gamma \) is 0.5 when the SR mechanism is not utilized, and 2.5 when it is, particularly at lower coverage levels. These findings highlight the importance of parameter tuning in achieving optimal performance and provide insights into the adaptability of the SMP scoring function under different operational settings. 


\subsection{Quantitative Results}
\begin{table}[!hbt]
\caption{Quantitative results for SelectiveNet and SYNC with coverage 0.7}
\label{table:SN-SYNC-Q}
\centering
\begin{tabular}{ccccc}
\toprule
\textbf{Prediction} & \multicolumn{2}{c}{\textbf{SelectiveNet}} & \multicolumn{2}{c}{\textbf{SYNC}} \\ 
\textbf{Sample} & \textbf{Correct} & \textbf{Incorrect} & \textbf{Correct} & \textbf{Incorrect} \\ \hline
Accept & 64.54\% & 5.44\% & 65.01\% & 4.99\% \\ 
Reject & 15.6\% & 14.42\% & 15.16\% & 14.84\% \\ 
\bottomrule
\end{tabular}

\end{table}

Table \ref{table:SN-SYNC-Q} shows the quantitative analysis on the ImageNet-100 dataset for the SYNC and SN approaches, respectively, for coverage 0.7. We observed that our model accepts more correctly classified samples while rejecting more incorrectly classified ones. Additionally, the percentage of accepted samples that are incorrectly predicted by the model and rejected samples that are correctly predicted has decreased.
\section{Conclusion}
In this work, we proposed a novel approach to enhance selective prediction in deep neural networks, focusing on improving accuracy and reliability under uncertainty. Building on prior work showing SR's effectiveness in enhancing selective methods, we incorporate SR directly into the training process. By integrating selective prior throughout training and inference phases, we promote a more holistic understanding and utilization of uncertainty. Our experimental results demonstrate the efficacy of our approach, improving selective prediction reliability and overall model performance across various datasets. This marks a step forward in developing robust and accurate AI systems, especially crucial for applications where precision is paramount. Our integrated approach represents a significant advancement, offering a foundation for future research in creating more dependable AI systems capable of navigating uncertainty effectively.

\bibliographystyle{IEEEtran}   
\bibliography{mybibfile} 

@article{zhang2021understanding,
  title={Understanding deep learning (still) requires rethinking generalization},
  author={Zhang, Chiyuan and Bengio, Samy and Hardt, Moritz and Recht, Benjamin and Vinyals, Oriol},
  journal={Communications of the ACM},
  volume={64},
  number={3},
  pages={107--115},
  year={2021},
  publisher={ACM New York, NY, USA}
}

@inproceedings{geifman2019selectivenet,
  title={SelectiveNet: A Deep Neural Network with an Integrated Reject Option},
  author={Geifman, Yonatan and El-Yaniv, Ran},
  booktitle={International Conference on Algorithmic Learning Theory},
  pages={67--82},
  year={2019},
  organization={Springer}
}

@inproceedings{
2,
title={EVALUATION OF NEURAL ARCHITECTURES TRAINED WITH SQUARE LOSS VS CROSS-ENTROPY IN CLASSIFICATION TASKS},
author={Like Hui and Mikhail Belkin},
booktitle={International Conference on Learning Representations},
year={2021},
}

@InProceedings{3,
  title = 	 {Trainable Calibration Measures for Neural Networks from Kernel Mean Embeddings},
  author =       {Kumar, Aviral and Sarawagi, Sunita and Jain, Ujjwal},
  booktitle = 	 {Proceedings of the 35th International Conference on Machine Learning},
  pages = 	 {2805--2814},
  year = 	 {2018},
  editor = 	 {Dy, Jennifer and Krause, Andreas},
  volume = 	 {80},
  series = 	 {Proceedings of Machine Learning Research},
  month = 	 {10--15 Jul},
  publisher =    {PMLR},
}

@inproceedings{5,
 author = {Mukhoti, Jishnu and Kulharia, Viveka and Sanyal, Amartya and Golodetz, Stuart and Torr, Philip and Dokania, Puneet},
 booktitle = {Advances in Neural Information Processing Systems},
 editor = {H. Larochelle and M. Ranzato and R. Hadsell and M.F. Balcan and H. Lin},
 pages = {15288--15299},
 publisher = {Curran Associates, Inc.},
 title = {Calibrating Deep Neural Networks using Focal Loss},
 volume = {33},
 year = {2020}
}

@inproceedings{10,
title={What Can we Learn From The Selective Prediction And Uncertainty Estimation Performance Of 523 Imagenet Classifiers?},
author={Ido Galil and Mohammed Dabbah and Ran El-Yaniv},
booktitle={The Eleventh International Conference on Learning Representations },
year={2023},
}

@article{el2010foundations,
  title={On the Foundations of Noise-free Selective Classification.},
  author={El-Yaniv, Ran and others},
  journal={Journal of Machine Learning Research},
  volume={11},
  number={5},
  year={2010}
}

@inproceedings{feng2022towards,
  title={Towards Better Selective Classification},
  author={Feng, Leo and Ahmed, Mohamed Osama and Hajimirsadeghi, Hossein and Abdi, Amir H},
  booktitle={The Eleventh International Conference on Learning Representations},
  year={2022}
}

@article{liu2019deep,
  title={Deep gamblers: Learning to abstain with portfolio theory},
  author={Liu, Ziyin and Wang, Zhikang and Liang, Paul Pu and Salakhutdinov, Russ R and Morency, Louis-Philippe and Ueda, Masahito},
  journal={Advances in Neural Information Processing Systems},
  volume={32},
  year={2019}
}

@article{geifman2017selective,
  title={Selective classification for deep neural networks},
  author={Geifman, Yonatan and El-Yaniv, Ran},
  journal={Advances in neural information processing systems},
  volume={30},
  year={2017}
}

@misc{gal2016dropout,
      title={Dropout as a Bayesian Approximation: Representing Model Uncertainty in Deep Learning}, 
      author={Yarin Gal and Zoubin Ghahramani},
      year={2016},
      eprint={1506.02142},
      archivePrefix={arXiv},
      primaryClass={stat.ML}
}

@inproceedings{he2016deep,
  title={Deep residual learning for image recognition},
  author={He, Kaiming and Zhang, Xiangyu and Ren, Shaoqing and Sun, Jian},
  booktitle={Proceedings of the IEEE conference on computer vision and pattern recognition (CVPR)},
  year={2016}
}

@inproceedings{zhang2018shufflenet,
  title={Shufflenet: An extremely efficient convolutional neural network for mobile devices},
  author={Zhang, Xiangyu and Zhou, Xinyu and Lin, Mengxiao and Sun, Jian},
  booktitle={Proceedings of the IEEE Conference on Computer Vision and Pattern Recognition (CVPR)},
  year={2018}
}

@article{holzinger2017we,
  title={What do we need to build explainable AI systems for the medical domain?},
  author={Holzinger, Andreas and Biemann, Chris and Pattichis, Constantinos S and Kell, Douglas B},
  journal={arXiv preprint arXiv:1712.09923},
  year={2017}
}

@misc{kirchenbauer2023watermark,
      title={A Watermark for Large Language Models}, 
      author={John Kirchenbauer and Jonas Geiping and Yuxin Wen and Jonathan Katz and Ian Miers and Tom Goldstein},
      year={2023},
      eprint={2301.10226},
      archivePrefix={arXiv},
      primaryClass={cs.LG}
}

@article{krizhevsky2009cifar,
  title={Learning multiple layers of features from tiny images},
  author={Krizhevsky, Alex and Hinton, Geoffrey},
  journal={Technical report, University of Toronto},
  year={2009}
}

@misc{stanford_cars_dataset,
  title = {Stanford Cars Dataset},
  author = {Fei-Fei Li and Justin Johnson and Li-Jia Li},
  year = {2013},
}

@article{imagenet_cvpr09,
  author    = {Jia Deng and
               Wei Dong and
               Richard Socher and
               Li{-}Jia Li and
               Kai Li and
               Li Fei{-}Fei},
  title     = {{ImageNet:} {A} Large-Scale Hierarchical Image Database},
  booktitle = {{CVPR} '09},
  year      = {2009},
  pages     = {248--255},
  doi       = {10.1109/CVPR.2009.5206848},
  note      = {ImageNet dataset, subset with 100 classes},
}

@inproceedings{
loshchilov2018decoupled,
title={Decoupled Weight Decay Regularization},
author={Ilya Loshchilov and Frank Hutter},
booktitle={International Conference on Learning Representations},
year={2019},
}

@article{cosine_annealing,
  author    = {Loshchilov, Ilya and Hutter, Frank},
  title     = {SGDR: Stochastic Gradient Descent with Warm Restarts},
  journal   = {arXiv preprint arXiv:1608.03983},
  year      = {2016},
}

@article{sgd,
  author    = {Leon Bottou},
  title     = {Stochastic Gradient Descent Tricks},
  journal   = {Neural Networks: Tricks of the Trade},
  year      = {2012},
  pages     = {421--436},
  publisher = {Springer},
}

@article{negative_entropy,
  author    = {Liu, Yang and Han, Wei and Gui, Jie and Han, Shuguang},
  title     = {Negative Entropy Regularization},
  journal   = {arXiv preprint arXiv:1907.04798},
  year      = {2019},
}

@article{paszke2019pytorch,
  title={Pytorch: An imperative style, high-performance deep learning library},
  author={Paszke, Adam and Gross, Sam and Massa, Francisco and Lerer, Adam and Bradbury, James and Chanan, Gregory and Killeen, Trevor and Lin, Zeming and Gimelshein, Natalia and Antiga, Luca and others},
  journal={Advances in neural information processing systems},
  volume={32},
  year={2019}
}

@misc{gatheluck_pytorch_selectivenet,
  title        = {{pytorch-SelectiveNet: Unofficial PyTorch implementation of "SelectiveNet: A Deep Neural Network with an Integrated Reject Option"}},
  author       = {gatheluck},
  year         = {2024},
  note         = {Accessed: 2024-02-09}
}

@article{tan2019efficientnet,
  title={EfficientNet: Rethinking Model Scaling for Convolutional Neural Networks},
  author={Tan, Mingxing and Le, Quoc V},
  journal={arXiv preprint arXiv:1905.11946},
  year={2019}
}

@article{bartlett2008classification,
  title={Classification with a Reject Option using a Hinge Loss.},
  author={Bartlett, Peter L and Wegkamp, Marten H},
  journal={Journal of Machine Learning Research},
  volume={9},
  number={8},
  year={2008}
}

@article{cortes2016boosting,
  title={Boosting with abstention},
  author={Cortes, Corinna and DeSalvo, Giulia and Mohri, Mehryar},
  journal={Advances in Neural Information Processing Systems},
  volume={29},
  year={2016}
}

@inproceedings{charoenphakdee2021classification,
  title={Classification with rejection based on cost-sensitive classification},
  author={Charoenphakdee, Nontawat and Cui, Zhenghang and Zhang, Yivan and Sugiyama, Masashi},
  booktitle={International Conference on Machine Learning},
  pages={1507--1517},
  year={2021},
  organization={PMLR}
}

@inproceedings{7344808,
  author={Loeffel, Pierre-Xavier and Marsala, Christophe and Detyniecki, Marcin},
  booktitle={2015 IEEE International Conference on Data Science and Advanced Analytics (DSAA)}, 
  title={Classification with a reject option under Concept Drift: The Droplets algorithm}, 
  year={2015},
  volume={},
  number={},
  pages={1-9},
  }

@article{JMLR:v24:21-0048,
  author  = {Vojtech Franc and Daniel Prusa and Vaclav Voracek},
  title   = {Optimal Strategies for Reject Option Classifiers},
  journal = {Journal of Machine Learning Research},
  year    = {2023},
  volume  = {24},
  number  = {11},
  pages   = {1--49},
  
}

@InProceedings{pmlr-v97-franc19a,
  title = 	 {On discriminative learning of prediction uncertainty},
  author =       {Franc, Vojtech and Prusa, Daniel},
  booktitle = 	 {Proceedings of the 36th International Conference on Machine Learning},
  pages = 	 {1963--1971},
  year = 	 {2019},
  editor = 	 {Chaudhuri, Kamalika and Salakhutdinov, Ruslan},
  volume = 	 {97},
  series = 	 {Proceedings of Machine Learning Research},
  month = 	 {09--15 Jun},
  publisher =    {PMLR},
  
}

@inproceedings{NIPS2011_4b6538a4,
 author = {Wiener, Yair and El-Yaniv, Ran},
 booktitle = {Advances in Neural Information Processing Systems},
 editor = {J. Shawe-Taylor and R. Zemel and P. Bartlett and F. Pereira and K.Q. Weinberger},
 pages = {},
 publisher = {Curran Associates, Inc.},
 title = {Agnostic Selective Classification},
 volume = {24},
 year = {2011}
}

@misc{mohri2023learning,
      title={Learning to Reject with a Fixed Predictor: Application to Decontextualization}, 
      author={Christopher Mohri and Daniel Andor and Eunsol Choi and Michael Collins},
      year={2023},
      eprint={2301.09044},
      archivePrefix={arXiv},
      primaryClass={cs.LG}
}

@inproceedings{NEURIPS2021_2cb6b103,
 author = {Granese, Federica and Romanelli, Marco and Gorla, Daniele and Palamidessi, Catuscia and Piantanida, Pablo},
 booktitle = {Advances in Neural Information Processing Systems},
 editor = {M. Ranzato and A. Beygelzimer and Y. Dauphin and P.S. Liang and J. Wortman Vaughan},
 pages = {5669--5681},
 publisher = {Curran Associates, Inc.},
 title = {DOCTOR: A Simple Method for Detecting Misclassification Errors},

 volume = {34},
 year = {2021}
}

@InProceedings{Xia_2022_ACCV,
    author    = {Xia, Guoxuan and Bouganis, Christos-Savvas},
    title     = {Augmenting Softmax Information for Selective Classification with Out-of-Distribution Data},
    booktitle = {Proceedings of the Asian Conference on Computer Vision (ACCV)},
    month     = {December},
    year      = {2022},
    pages     = {1995-2012}
}

\end{document}


\title{Supplementary Material}

\markboth{Journal of IEEE Transactions on Artificial Intelligence, Vol. 00, No. 0, Month 2020}
{ \MakeLowercase{\textit{et al.}}: Selective Prior Synchronization via SYNC Loss}

\maketitle

\section{Lipschitz‐based Upper Bound on the SMP Hyper–parameter}
\label{app:lipschitz_bound}

\subsection{Proof of Lemma 1}
Let $u,v\in\Delta^{C-1}$ and 
\[
a:=m(u):=\max_i u_i,\qquad b:=m(v):=\max_i v_i.
\]
We proceed in three steps.

\medskip
\noindent\textbf{Step 1: Range of the maxima.}
Since $u$ and $v$ are probability vectors ($u_i\ge 0$, $\sum_i u_i=1$), each coordinate is at most $1$, hence $a,b\le 1$.
Moreover, at least one coordinate must be at least the average $1/C$, hence $a,b\ge 1/C$.
Thus
\begin{equation}
\label{eq:ab-range}
a,b\in\bigl[\,1/C,\,1\,\bigr].
\end{equation}

\medskip
\noindent\textbf{Step 2: The maximum–coordinate function is $1$-Lipschitz in $\ell_\infty$.}
Define $M:\mathbb{R}^C\!\to\!\mathbb{R}$ by $M(u):=\max_{1\le i\le C} u_i$.
We claim that for any $u,v\in\mathbb{R}^C$,
\begin{equation}
\label{eq:max-1lip}
|M(u)-M(v)| \;=\; \bigl|\,\max_i u_i - \max_i v_i\,\bigr| \;\le\; \|u-v\|_\infty.
\end{equation}
Indeed,
\[
\begin{aligned}
\max_i u_i - \max_i v_i
&\le \max_i (u_i - v_i)\\
&\le \max_i |u_i - v_i|\\
&= \|u-v\|_\infty.
\end{aligned}
\]
and swapping $(u,v)$ yields $\max_i v_i - \max_i u_i \le \|u-v\|_\infty$.
Combining the two inequalities gives \eqref{eq:max-1lip}.

\medskip
\noindent\textbf{Step 3: Apply the Mean Value Theorem to $t^\gamma$.}
Let $f(t):=t^\gamma$ with $\gamma>0$.
By \eqref{eq:ab-range}, $a,b\in[1/C,1]\subset(0,\infty)$, so $f$ is continuous on $[b,a]$ and differentiable on $(b,a)$, and the Mean Value Theorem applies:
there exists $\xi\in[b,a]$ such that
\begin{equation}
\label{eq:MVT}
a^\gamma - b^\gamma \;=\; f(a)-f(b) \;=\; f'(\xi)\,(a-b)
\;=\; \gamma\,\xi^{\,\gamma-1}\,(a-b).
\end{equation}
Taking absolute values in \eqref{eq:MVT} and using \eqref{eq:max-1lip} gives
\begin{equation}
\label{eq:abs-core}
|a^\gamma - b^\gamma|
\;=\; \gamma\,\xi^{\,\gamma-1}\,|a-b|
\;\le\; \gamma\,\xi^{\,\gamma-1}\,\|u-v\|_\infty.
\end{equation}
It remains to bound $\xi^{\gamma-1}$ using that $\xi\in[1/C,1]$ (since $\xi\in[b,a]$ and $a,b\in[1/C,1]$).

\smallskip
\emph{Case $\gamma\ge 1$.}
Then $\gamma-1\ge 0$ and $t\mapsto t^{\gamma-1}$ is nondecreasing on $[0,\infty)$.
Hence $\xi^{\gamma-1}\le 1^{\gamma-1}=1$ for all $\xi\le 1$.
Plugging this into \eqref{eq:abs-core} yields
\[
|a^\gamma - b^\gamma|
\;\le\; \gamma\,\|u-v\|_\infty.
\]

\smallskip
\emph{Case $0<\gamma<1$.}
Then $\gamma-1<0$ and $t\mapsto t^{\gamma-1}$ is nonincreasing on $(0,\infty)$.
Using $\xi\ge 1/C$ we obtain
\[
\xi^{\gamma-1} \;\le\; \bigl(1/C\bigr)^{\gamma-1} \;=\; C^{\,1-\gamma}.
\]
Substituting into \eqref{eq:abs-core} gives
\[
|a^\gamma - b^\gamma|
\;\le\; \gamma\,C^{\,1-\gamma}\,\|u-v\|_\infty.
\]

\medskip
Finally, note that $s_\gamma(u)=a^\gamma$ and $s_\gamma(v)=b^\gamma$, so
\[
\bigl|s_\gamma(u)-s_\gamma(v)\bigr|
\;=\; |a^\gamma-b^\gamma|
\;\le\;
\begin{cases}
\gamma\,\|u-v\|_\infty, & \gamma\ge 1,\\[4pt]
\gamma\,C^{\,1-\gamma}\,\|u-v\|_\infty, & 0<\gamma<1,
\end{cases}
\]
which is the stated Lipschitz bound.

\subsection{Proof of Corollary 1}

Let
\[
\mathcal F_\gamma:\; x \;\longmapsto\;
s_\gamma\!\bigl(\mathrm{softmax}\bigl(z(x)\bigr)\bigr),
\]
where $z:\mathbb R^{d}\!\to\!\mathbb R^{C}$ denotes the network logits,
$\mathrm{softmax}$ acts component–wise on $z(x)$ to give probabilities
$p(x)$, and $s_\gamma$ is the \emph{Softmax-Power (SMP) score}
defined in Lemma 1 in the manuscript.
Our goal is to ensure that the \emph{entire} mapping $\mathcal F_\gamma$
is $1$–Lipschitz with respect to the input, i.e.
\begin{equation}\label{eq:one_lip}
    \bigl|\,s_\gamma\bigl(p(x_1)\bigr)-
           s_\gamma\bigl(p(x_2)\bigr)\bigr|
    \;\le\;
    \|x_1-x_2\|_2
    \quad\forall\,x_1,x_2\in\mathbb R^{d}.
\end{equation}

Because Lipschitz moduli \emph{multiply} under composition,
condition~\eqref{eq:one_lip} is satisfied whenever
\begin{equation}
    \underbrace{\bigl\|\nabla_x z(x)\bigr\|_2}_{\text{backbone}}
    \;\times\;
    \underbrace{\lambda_{\max}\!\bigl(\nabla_z\,\mathrm{softmax}(z)\bigr)}_
               {\text{soft-max}}
    \;\times\;
    \underbrace{L_\gamma}_{\text{SMP (Lemma 1)}}
    \;\le\;1.
    \label{eq:prod_rule}
\end{equation}
Below we bound each factor in~\eqref{eq:prod_rule}.

\subsubsection{Backbone factor}

\begin{equation}
    L_z\;:=\;\bigl\|\nabla_x z(x)\bigr\|_2
           \quad(\text{data-dependent}).
    \label{eq:Lz_def}
\end{equation}
In practice we measure $L_z=\max_j\|\nabla_x z(x_j)\|_2$ on the current
mini-batch.


\subsubsection{Softmax factor}
The Jacobian of the softmax is $J=\nabla_z\,\mathrm{softmax}(z)=\mathrm{diag}(p)-pp^\top$.
Each row/column absolute sum is $2p_i(1-p_i)\le \tfrac12$, hence
$\lambda_{\max}\!\bigl(\nabla_z\,\mathrm{softmax}(z)\bigr)=\|J\|_2\le \tfrac12.$

\noindent\emph{Optional (temperature).} If $p=\mathrm{softmax}(z/\tau)$, then
$\nabla_z\,\mathrm{softmax}(z/\tau)=\tfrac1{\tau}J(z/\tau)$ and
$\lambda_{\max}\!\bigl(\nabla_z\,\mathrm{softmax}(z/\tau)\bigr)\le \tfrac{1}{2\tau}$.
Setting $B:=2\tau$ recovers the shorthand $\lambda_{\max}\le 1/B$.

As per our assumption in section C, logits are bounded as
\begin{equation}
    \|z(x)\|_\infty\;\le\;B.
    \label{eq:logit_bound}
\end{equation}
The Jacobian of the soft-max has entries
$\partial p_i / \partial z_j = p_i(\delta_{ij}-p_j)$, hence
$|\partial p_i/\partial z_j|
    \le\max_{t\in[0,1]}t(1-t)=\tfrac14$.
Therefore
\[
    \lambda_{\max}\!\bigl(\nabla_z\,\mathrm{softmax}(z)\bigr)
    \;\le\;\tfrac14.
\]
A slightly looser bound that exposes the role of $B$ and is valid for
all $B\ge1$ is
\begin{equation}
    \lambda_{\max}\!\bigl(\nabla_z\,\mathrm{softmax}(z)\bigr)
    \;\le\;\frac1B,
    \qquad(B\ge1).
    \label{eq:softmax_bound}
\end{equation}

\subsubsection{SMP factor}

By Lemma 1, the SMP score is Lipschitz on
$\Delta^{C-1}$ with modulus
\[
L_\gamma \;=\;
\begin{cases}
\gamma, & \gamma \ge 1,\\[2pt]
\gamma\,C^{\,1-\gamma}, & 0<\gamma<1.
\end{cases}
\]
Thus, for any $u,v\in\Delta^{C-1}$,
\begin{equation}
    \bigl|s_\gamma(u)-s_\gamma(v)\bigr|
    \;\le\;
    L_\gamma\,\|u-v\|_\infty.
    \label{eq:smp_lip_piecewise}
\end{equation}

\subsubsection{Combining the bounds}

Substituting~\eqref{eq:Lz_def}, \eqref{eq:softmax_bound},
and~\eqref{eq:smp_lip_piecewise} into~\eqref{eq:prod_rule} yields
\[
    L_z \times \frac{1}{B} \times L_\gamma \;\le\; 1.
\]
Equivalently, the SMP exponent must satisfy the piecewise constraint
\begin{equation}
\boxed{
\begin{aligned}
\gamma \;&\le\; \frac{B}{L_z} \;=\; \frac{B}{\|\nabla_x z(x)\|_2},
&& \text{if }\gamma \ge 1,\\[6pt]
\gamma\,C^{\,1-\gamma} \;&\le\; \frac{B}{L_z} \;=\; \frac{B}{\|\nabla_x z(x)\|_2},
&& \text{if }0<\gamma<1.
\end{aligned}}
\label{eq:gamma_bound_piecewise_appendix}
\end{equation}

\subsubsection{Interpretation}

\paragraph{Sharpness–stability trade-off.}
A larger $\gamma$ sharpens the SMP score—improving separation between
``easy'' and ``uncertain'' examples—but simultaneously increases the
overall Lipschitz constant of the network.  Inequality~\eqref{eq:prod_rule}
quantifies the permissible range.

\paragraph{Data-driven selection.}
Compute $L_z$ on a representative mini-batch, plug in the empirical logit
bound $B$ (typically $B\approx10$), and pick any
$\gamma$ that respects the constraint.
In our experiments $L_z\in[4,20]$ implies $\gamma\lesssim2.5$,
which is exactly the range used in our ablation study.

\begin{figure}[h]
  \centering
  \begin{subfigure}[b]{0.3\textwidth}
    \includegraphics[width=\textwidth]{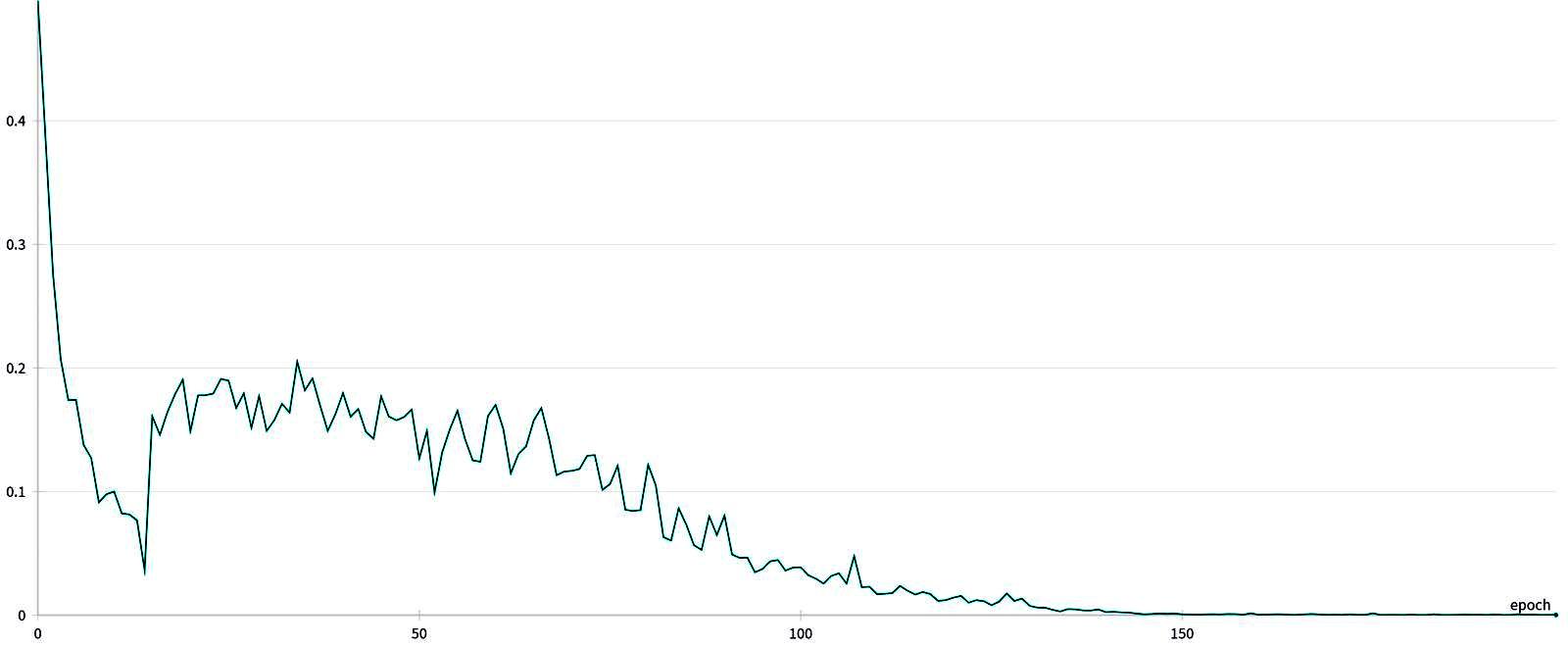}
    \caption{SMP with $\gamma=0.5$}
    \label{fig:image1}
    \vspace{0.5cm}
  \end{subfigure}
  \hfill
  \begin{subfigure}[b]{0.3\textwidth}
    \includegraphics[width=\textwidth]{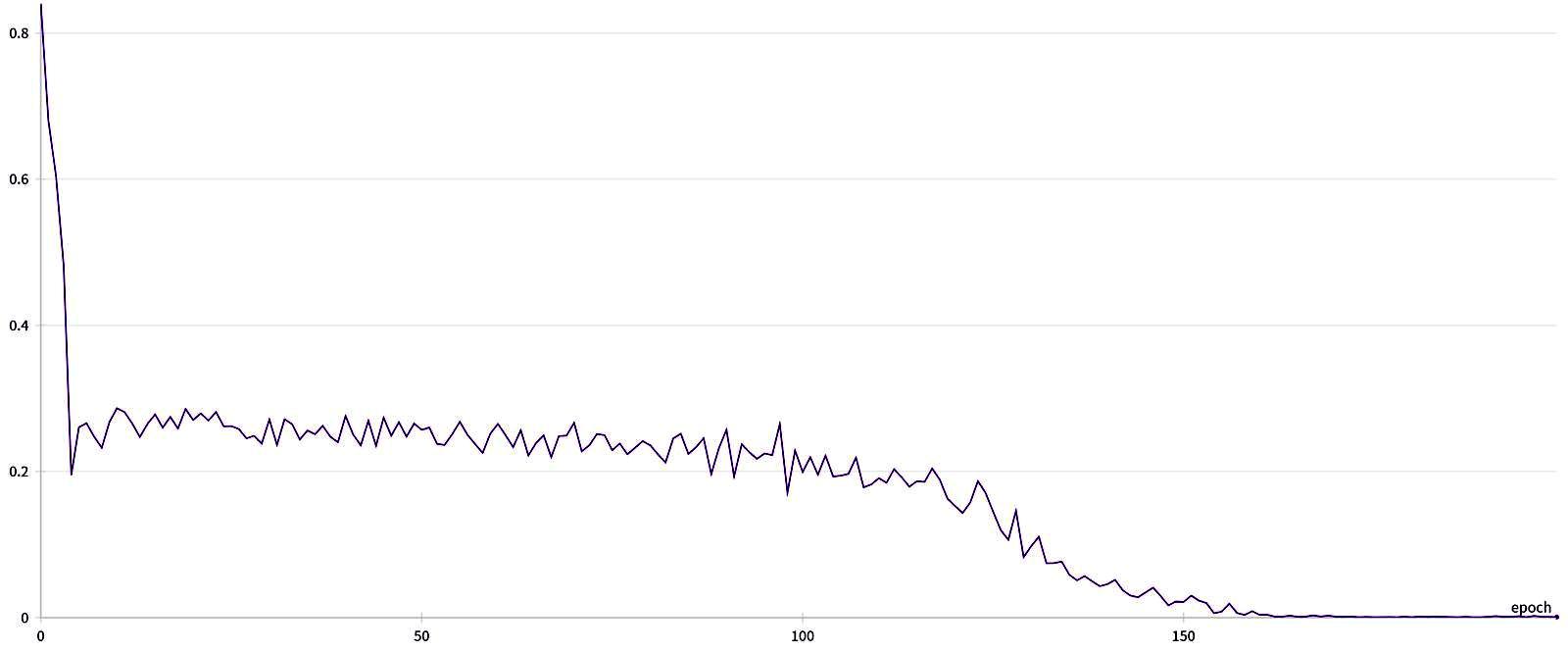}
    \caption{SMP with $\gamma=1$}
    \label{fig:image2}
    \vspace{0.5cm}
  \end{subfigure}
  \hfill
  \begin{subfigure}[b]{0.3\textwidth}
    \includegraphics[width=\textwidth]{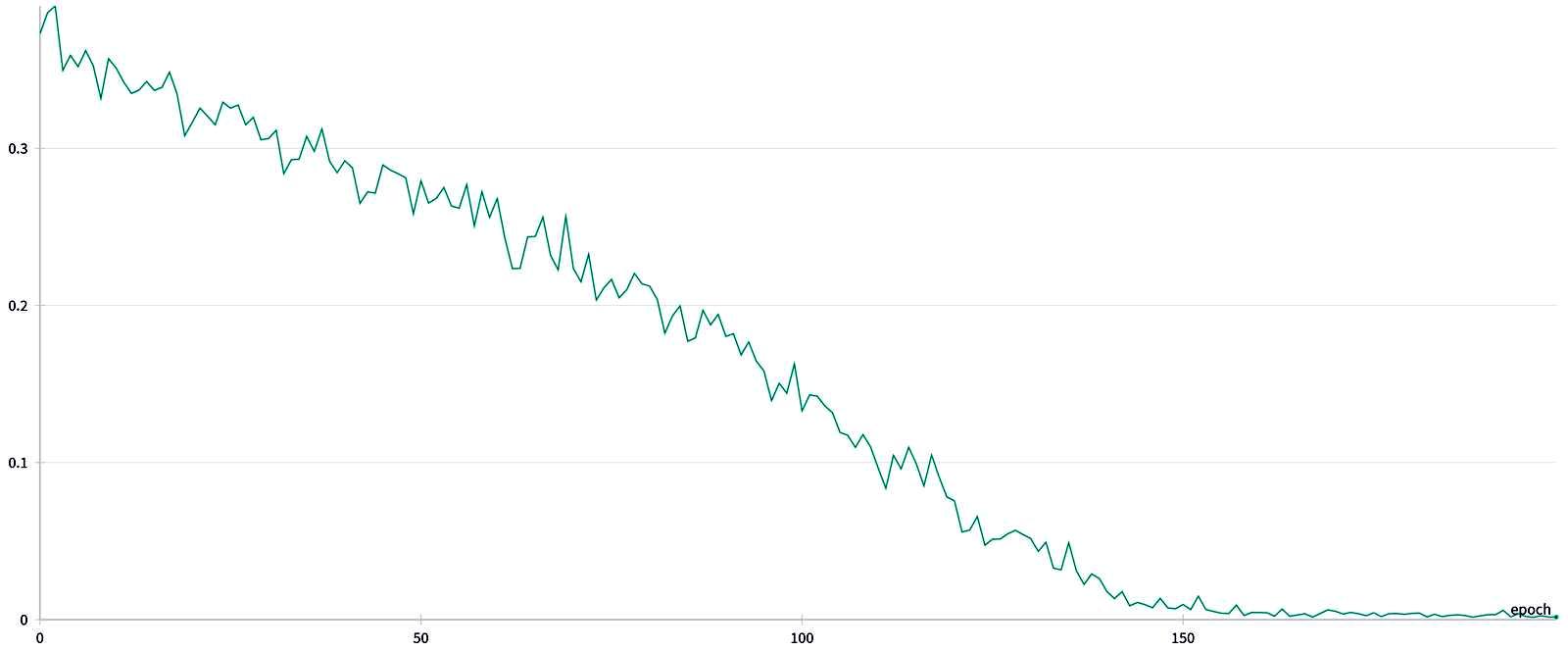}
    \caption{SMP with $\gamma=2.5$}
    \label{fig:image3}
    \vspace{0.5cm}
  \end{subfigure}
  \caption{Convergence analysis of our sync loss with different SMP values.}
  \label{fig:convergence_plotss}
  \vspace{0.5cm}
\end{figure}

\subsection{Proof of Proposition }
\begin{proof}
We treat $x$ as fixed throughout and abbreviate
$g\!:=g_\theta(x)\in[0,1]$ and $s\!:=s_\gamma\!\bigl(p(x)\bigr)\in[0,1]$.

\paragraph{Hessian of the SYNC quadratic term.}
Define
\(
h(\theta) \;=\; \mu\,(g-s)^{2}.
\)
A direct differentiation gives
\[
\nabla_{\theta}h
    = 2\mu\,(g-s)\nabla_{\theta}g,
\qquad
\nabla_{\theta}^{2}h
    =
    2\mu\,
    \Bigl[
        \nabla_{\theta}g\,\nabla_{\theta}g^{\!\top}
      + (g-s)\,\nabla_{\theta}^{2}g
    \Bigr].
\]

\paragraph{Bounding the first summand.}
By hypothesis H-2,
\(
\|\nabla_{\theta}g\|_{2}\le G_{\!*}.
\)
Hence
\[
\bigl\|\nabla_{\theta}g\,\nabla_{\theta}g^{\!\top}\bigr\|_{2}
  = \|\nabla_{\theta}g\|_{2}^{2}
  \;\le\; G_{\!*}^{2}.
\]

\paragraph{Bounding the second summand.}
Because $g,s\in[0,1]$,
\(|g-s|\le 1\).
For scalar network outputs,
\(
\|\nabla_{\theta}^{2}g\|_{2}
  \le
  \|\nabla_{\theta}g\|_{F}^{2}
  \le
  G_{\!*}^{2},
\)
using the inequality
\(\|\mathbf a\mathbf a^{\!\top}\|_{2}=\|\mathbf a\|_{2}^{2}\).
Therefore
\[
\bigl\|(g-s)\,\nabla_{\theta}^{2}g\bigr\|_{2}
  \;\le\; G_{\!*}^{2}.
\]

\paragraph{Aggregating the bounds.}
Combining Steps~2 and~3 yields
\[
\bigl\|\nabla_{\theta}^{2}h\bigr\|_{2}
  \;\le\;
  2\mu\,G_{\!*}^{2}.
\]

\paragraph{Adding the baseline Hessian.}
By hypothesis H1 from the manuscript,
\(\|\nabla_{\theta}^{2}\mathcal L_{\mathrm{SN}}\|_{2}\le L.\)
Hence, for any $\theta$ and $x$,
\[
\bigl\|\nabla_{\theta}^{2}\mathcal L_{\mathrm{SYNC}}\bigr\|_{2}
  \;\le\;
  L + 2\mu G_{\!*}^{2},
\]
which is the matrix inequality equation (21) in the manuscript.  \qedhere
\end{proof}
\section{Appendix}\label{appendix}

\subsection{Datasets}\label{Datasets}


\textbf{CIFAR-100} \cite{krizhevsky2009cifar}: The CIFAR-100 dataset is a widely used benchmark in computer vision for image recognition tasks. CIFAR-100 consists of 60,000 32$\times$32 RGB images in 100 classes, each class having 600 images, offering a more challenging task due to its finer classification granularity.

\textbf{ImageNet-100}: ImageNet-100 is a subset of the ImageNet dataset~\cite{imagenet_cvpr09}, which is widely recognized in the field of computer vision and machine learning for its extensive collection of annotated images. Designed for benchmarking and training deep learning models. The ImageNet-100 subset contains image-label pairs from 100 classes, the training set contains 1300 images for each class and 50 images for each class in the test set.

\textbf{Stanford Cars Dataset}~\cite{stanford_cars_dataset}:
The Stanford Cars dataset is a comprehensive collection of images specifically designed for fine-grained categorization and automatic recognition of car models. It contains over 16,000 images spanning 196 classes of car images categorized by car models. The dataset is divided into a training set, which comprises approximately 8,144 images, and a test set with around 8,041 images.




\begin{figure*}[!hbt]
    \centering
    \includegraphics[scale=0.2]{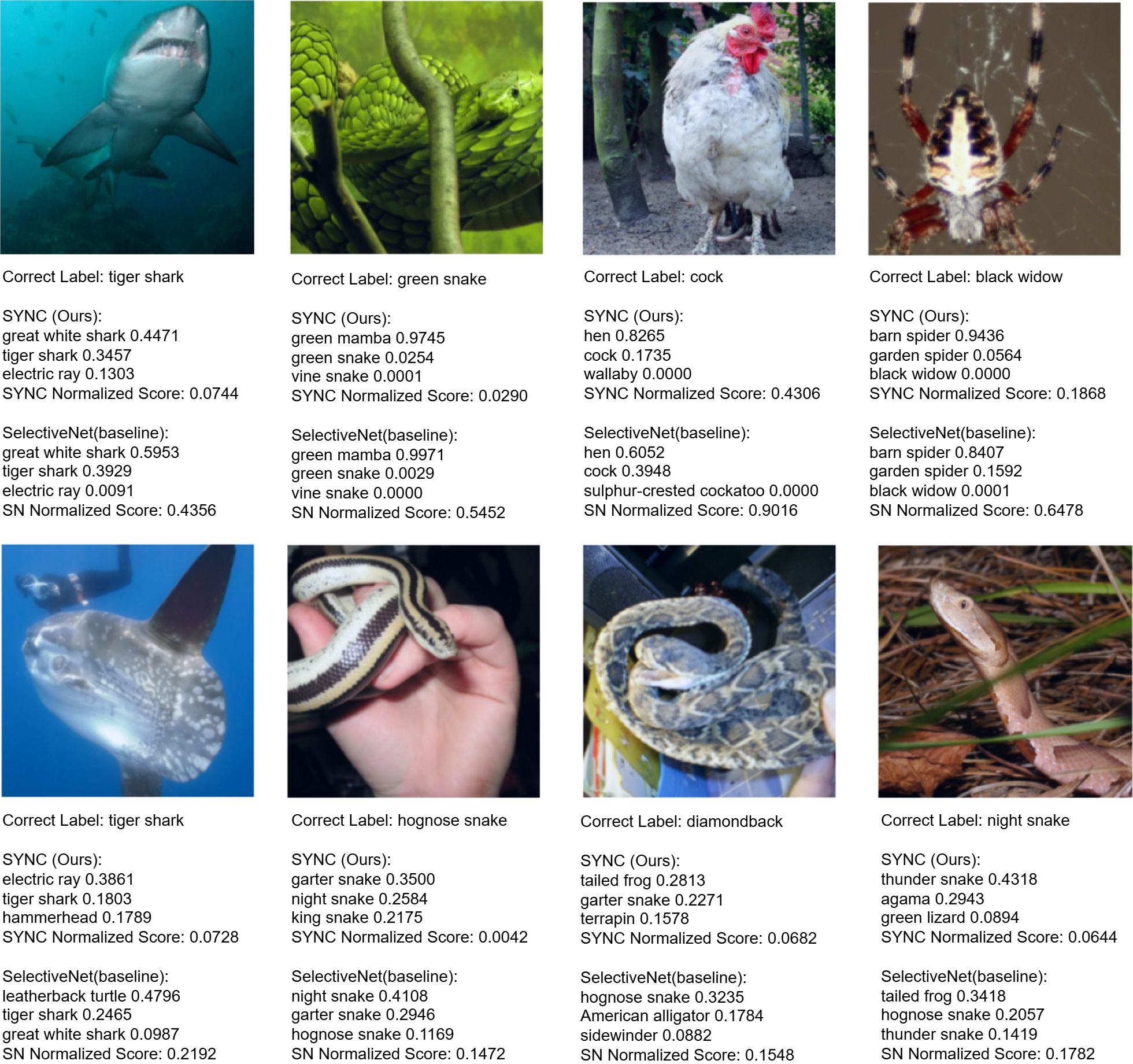}
    \vspace{0.5cm}
    \caption{It shows the image with its corresponding ground truth and top three logits given by the model using SYNC (our) approach and SelectiveNet (baseline) on the ImageNet dataset.}
    \vspace{0.5cm}
    \label{fig:add_eg}
    
\end{figure*}
\subsection{Experiment Settings}\label{Settings}

\textbf{Architectures}
In our implementation of DG and the core block of SN, we utilize a ResNet-18 \cite{he2016deep} architecture, applied across three datasets: CIFAR-100 \cite{krizhevsky2009cifar}, Stanford Cars \cite{stanford_cars_dataset}, and ImageNet-100. We replace the first 7$\times$7 max pooling layer with a 3$\times$3 pooling layer for CIFAR-100. For SN and SYNC, following the public implementation~\citep{gatheluck_pytorch_selectivenet}, we adopted a selective head consisting of one MLP layer with a sigmoid layer to get a selective score in the range 0 to 1. 

We utilize a ResNet-18 architecture in our implementation of Deep Gamblers and the core block of SelectiveNet, applied across CIFAR-100, Stanford Cars, and ImageNet100 datasets. The selective head of SelectiveNet consisits of a MLP layer with 512 hidden dimension, and a sigmoid layer to normalize the output to the range of 0 and 1 for selective score prediction.
end


\textbf{Hyperparameters} For CIFAR-100 and Stanford Cars datasets, we use AdamW \citep{loshchilov2018decoupled} optimizer with $\beta_1$= 0.9, $\beta_2$ = 0.99, weight decay as 0.05, a learning rate of 0.001 and trained for 90 epochs. For ImageNet-100 dataset, we use the SGD \citep{sgd} optimizer with a learning rate of 0.01 and weight decay of $5 \times 10^{-4}$. We apply cosine annealing \citep{cosine_annealing} scheduler for all experiments. The batch size is set to 64. We follow the data augmentation settings as in ~\cite{feng2022towards}. 
The SN hyperparameters $\lambda$ and $\alpha$ are 6 and 0.5 respectively for all the experiments. Our sync loss hyperparameter, $\mu$ is $1.0$ for all experiments. The score function hyperparameter, $\gamma$, is varied for different datasets and selection mechanisms. We tuned the training coverage of SN and SYNC from $10$ to $100$. We use 6, 30, and 50 as the reward parameters of the DG loss for CIFAR-100, Stanford Cars, and ImageNet-100, respectively. 
We use different values of $\alpha$ for different datasets because $\alpha$ affects the shape of the distribution curve of SR values. For datasets such as ImageNet-100, where we lack optimal hyperparameters from the respective baseline authors, we ensure a fair comparison by using the same hyperparameters across all baselines. For instance, our approach also employs the SN loss; the hyperparameters for both the SN and SYNC experiments are kept consistent to demonstrate the improvement over SN. For DG, we utilize the hyperparameters recommended by the authors for various datasets. The hyperparameters $\lambda$ and $\alpha$ are originally the SN hyperparameters. We introduced only $\mu$ and $\gamma$ hyperparameters, where $\mu$ is the loss balancing factor and $\gamma$ is SMP hyperparameter.

Our experiments are implemented in PyTorch~\citep{paszke2019pytorch}. The implementation of DG is based on the official code~\cite{feng2022towards, liu2019deep}. The SN implementation is based on the public code base ~\citep{gatheluck_pytorch_selectivenet}.
\subsection{Empirical Convergence of the SYNC Objective}
\label{sec:convergence}

\paragraph{Theoretical expectation.} 
Proposition 2 in the manuscript shows that
the SYNC objective is globally $(L+2\mu G_{\!*}^{2})$–smooth.
Hence stochastic gradient descent with any fixed stepsize
$\eta\le(L+2\mu G_{\!*}^{2})^{-1}$ enjoys the standard
$O(1/\!\sqrt{T})$ convergence rate for non-convex objectives.
Because $\mu$ is fixed at~$1$ throughout all experiments, the
smoothness multiplier increases only through the selector Jacobian
and \emph{linearly} with the SMP exponent~$\gamma$
(cf.\ Lemma 1 in the manuscript).
Our empirical study therefore tests whether larger values of~$\gamma$
lead merely to a proportional slowdown—as the theory predicts—while
preserving monotone descent and bounded gradients.

\paragraph{Experimental setup.}
We train the selector on CIFAR-100 with three exponents
$\gamma\in\{0.5,\,1,\,2.5\}$ and record the SYNC loss at every iteration.
The backbone and optimiser settings are \emph{identical} across runs,
so differences in the curves reflect solely the change in~$\gamma$.
Convergence trajectories are displayed in
Fig.~\ref{fig:convergence_plotss}.

\paragraph{Observations.}
\begin{itemize}[leftmargin=15pt]
    \item \textbf{Monotone descent and stability.}\;
          All three curves decrease smoothly without oscillation,
          consistent with the global smoothness guarantee of
          Proposition 2 in the manuscript.
          No gradient explosions were observed.
    \item \textbf{Effect of $\gamma$.}\;
          Higher $\gamma$ produces a visibly larger Lipschitz constant
          (steeper upper bound), so the step size $\eta$ becomes
          effectively more conservative.
          This manifests as the mild slowdown from
          $\gamma\!=\!0.5$ to $\gamma\!=\!2.5$, yet the final loss
          values coincide after $\approx 12\,000$ iterations,
          confirming that smoothness—not curvature pathologies—
          dictates the speed gap.
    \item \textbf{Alignment dynamics.}\;
          The additional gradients,
          $\Delta_f
          $ and
          $\Delta_g
          $, 
          are bounded by construction,
          preserving the stability of the baseline
          $\nabla_\theta\mathcal L_{\mathrm{SN}}$.
        
\end{itemize}

The empirical curves fully agree with the analytical bound
(equation 21 in the manuscript): adding the quadratic SYNC regulariser
merely rescales the smoothness constant,
preserves gradient stability,
and leads to consistent convergence across all tested values
of the SMP exponent~$\gamma$.

\subsection{Additional examples}\label{ad_ex}
The additional examples are shown in Fig. \ref{fig:add_eg}. We observe that our approach calibrates the rejection logit score and the softmax score to improve the selective classification ability of the model irrespective of the selection mechanism, i.e., with or without SR.

\bibliographystyle{IEEEtran}   
\bibliography{mybibfile}